\documentclass{article}

\usepackage{microtype}
\usepackage{graphicx}
\usepackage{booktabs} 

\usepackage{subcaption}
\usepackage{multirow}
\usepackage{makecell}
\usepackage{capt-of}
\usepackage{longtable}

\usepackage{hyperref}

\usepackage[accepted]{icml2025}  

\usepackage{amsmath}
\usepackage{amssymb}
\usepackage{mathtools}
\usepackage{amsthm}
\usepackage{array}
\usepackage[table]{xcolor}
\usepackage{colortbl}
\usepackage{xcolor}
\usepackage{enumitem}
\usepackage{fontawesome5}

\newcommand{\few}{\textcolor{orange}{\textbf{Few}}}
\newcommand{\full}{\textcolor{gray}{Full}}
\newcommand{\mcell}[2][c]{\begin{tabular}{@{}#1@{}}#2\end{tabular}}

\newcommand{\redbox}{%
  \begingroup%
  \setlength{\fboxsep}{0pt}%
  \setlength{\fboxrule}{1pt}%
  \fcolorbox{red}{white}{%
    \rule{0pt}{5pt}%
    \rule{5pt}{0pt}%
  }%
  \endgroup%
}

\usepackage[capitalize,noabbrev]{cleveref}

\theoremstyle{plain}
\newtheorem{theorem}{Theorem}[section]

\theoremstyle{definition}
\newtheorem{definition}[theorem]{Definition}

\theoremstyle{remark}

\usepackage[textsize=tiny]{todonotes}

\icmltitlerunning{MOTIF: Learning Action Motifs for Few-shot Cross-Embodiment Transfer}

\begin{document}

\twocolumn[
\icmltitle{MOTIF: Learning Action Motifs for Few-shot Cross-Embodiment Transfer}



\icmlsetsymbol{equal}{*}
\icmlsetsymbol{corr}{\faEnvelope}

\begin{icmlauthorlist}
\icmlauthor{Heng Zhi}{Tongji,equal}
\icmlauthor{Wentao Tan}{Tongji,equal}
\icmlauthor{Lei Zhu}{Tongji,corr}
\icmlauthor{Fengling Li}{UTS}
\icmlauthor{Jingjing Li}{UESTC}
\icmlauthor{Guoli Yang}{AIBD}
\icmlauthor{Heng Tao Shen}{Tongji}
\end{icmlauthorlist}

\icmlaffiliation{Tongji}{Tongji University}
\icmlaffiliation{UTS}{University of Technology Sydney}
\icmlaffiliation{UESTC}{University of Electronic Science and Technology of China}
\icmlaffiliation{AIBD}{Advanced Institute of Big Data}

\icmlcorrespondingauthor{Lei Zhu}{leizhu0608@gmail.com}

\icmlkeywords{Machine Learning, Robot Learning}

\vskip 0.3in
]



\printAffiliationsAndNotice{\icmlEqualContribution} 

\begin{abstract} 
While vision-language-action (VLA) models have advanced generalist robotic learning, cross-embodiment transfer remains challenging due to kinematic heterogeneity and the high cost of collecting sufficient real-world demonstrations to support fine-tuning. Existing cross-embodiment policies typically rely on shared-private architectures, which suffer from limited capacity of private parameters and lack explicit adaptation mechanisms. To address these limitations, we introduce MOTIF for efficient few-shot cross-embodiment transfer that decouples embodiment-agnostic spatiotemporal patterns, termed \textbf{action motifs}, from heterogeneous action data. Specifically, MOTIF first learns unified motifs via vector quantization with progress-aware alignment and embodiment adversarial constraints to ensure temporal and cross-embodiment consistency. We then design a lightweight predictor that predicts these motifs from real-time inputs to guide a flow-matching policy, fusing them with robot-specific states to enable action generation on new embodiments. Evaluations across both simulation and real-world environments validate the superiority of MOTIF, which significantly outperforms strong baselines in few-shot transfer scenarios by 6.5\% in simulation and 43.7\% in real-world settings. Code is available at \url{https://github.com/buduz/MOTIF}.
\end{abstract}

\begin{figure}[t!]
  \vskip 0.2in
  \begin{center}
    \centerline{\includegraphics[width=\columnwidth]{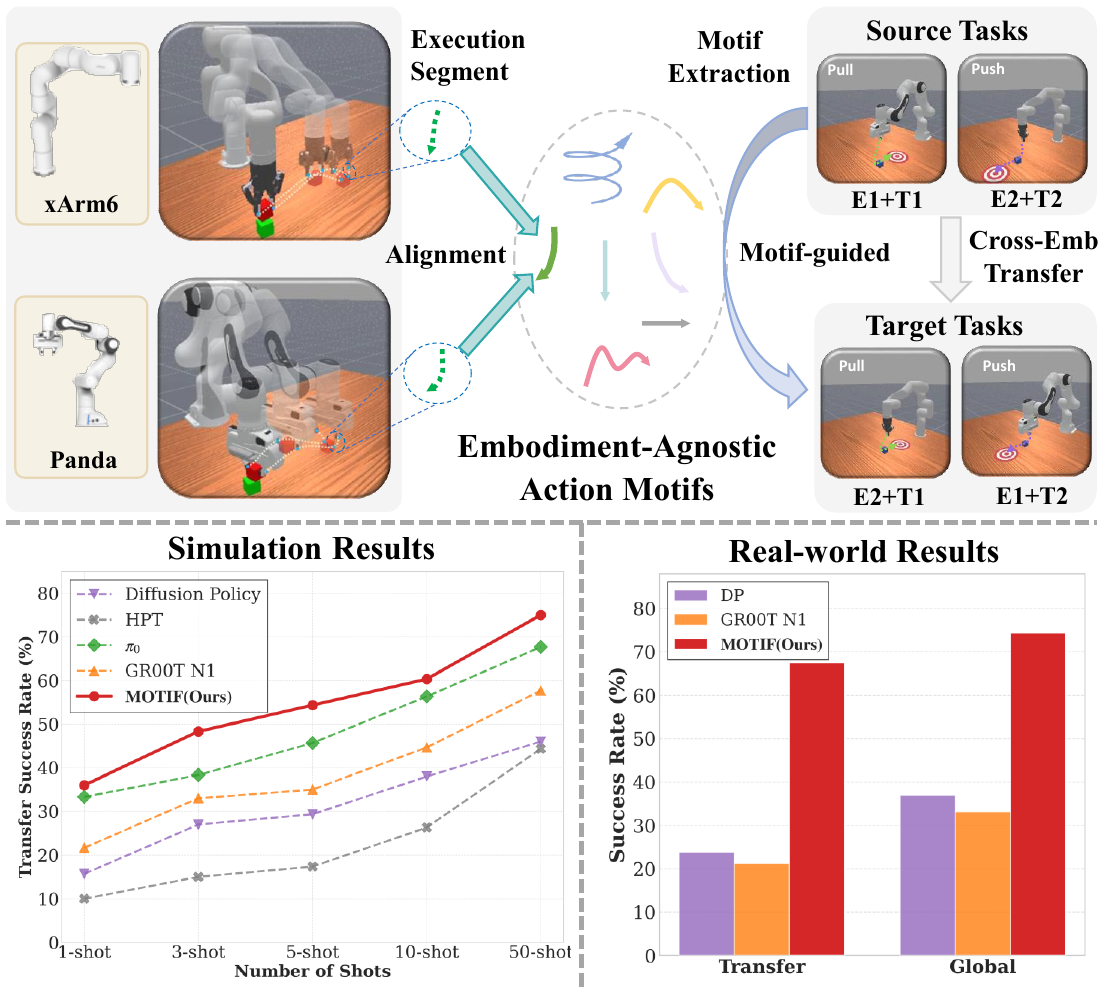}}
    \caption{\textbf{Concept and Performance of MOTIF.} \textbf{(Top) Motif-guided Transfer.} MOTIF extracts embodiment-agnostic \textit{action motifs} by aligning execution segments from heterogeneous robots (e.g., xArm6, Panda), bridging kinematic gaps for cross-embodiment transfer. The schematic illustrates how task behaviors learned by a source embodiment $E1$ on task $T1$ are adapted to a target ($E2+T1$). \textbf{(Bottom Left) Simulation Results.} MOTIF consistently outperforms strong baselines in Transfer Success Rate across all data regimes (1- to 50-shot). \textbf{(Bottom Right) Real-world Results.} Physical evaluations further validate this effectiveness, demonstrating significant improvements in both Transfer and Global success rates against SOTA methods.}
    \label{fig:teaser}
  \end{center}
  \vspace{-8mm}
\end{figure}

\section{Introduction} 
Embodied AI~\cite{embodied_survey_1,embodied_survey_2,embodied_survey_3} research aims to build generalist agents capable of perceiving and interacting with complex physical environments. Recently, driven by progress in multimodal large language models (MLLMs)~\cite{llava,Gpt4,Qwen2_5_vl}, robotic learning is transitioning from specialized visual policies~\cite{ACT,MDT,Diffusion-Policy} to general-purpose vision-language-action models (VLAs)~\cite{Openvla,pi_0} that inherit the semantic reasoning and world knowledge of MLLMs. By pre-training on internet-scale data across heterogeneous embodiments~\cite{RT-X}, VLAs acquire generalizable manipulation priors and physical commonsense. This paradigm integrates perception, reasoning, and control into a unified framework that grounds high-level semantics directly into low-level actions, facilitating zero-shot adaptation to novel tasks and open-world settings.

Despite internet-scale pre-training, transferring VLAs to new robots is hindered by two critical challenges:
(1) \textit{\textbf{Cross-Embodiment Misalignment.}} Kinematic heterogeneity leads to significant differences in action spaces that make source policies physically infeasible on the target embodiments, thereby restricting direct transfer.
(2) \textit{\textbf{Data Scarcity in New Embodiments.}} While fine-tuning effectively reduces domain shifts, the high cost of data acquisition in new embodiments often makes collecting sufficient demonstrations impractical. This scarcity forces models to rely on few-shot learning, which is often insufficient for generalizing to complex and unseen scenarios.

To achieve efficient cross-embodiment transfer, recent methods such as HPT~\cite{HPT} and GR00T N1~\cite{GR00T-N1} extend the vision-language-action (VLA) paradigm by designing shared-private architectures. Although these frameworks adapt to diverse embodiments via frozen backbones and fine-tuned private modules, their efficacy is constrained by two critical limitations:
(1) \textit{\textbf{Restricted Private Parameter Capacity.}} The limited capacity of the embodiment-specific private parameters impairs the alignment of heterogeneous action and state spaces within the shared embedding manifold.
(2) \textit{\textbf{Absence of Explicit Transfer Mechanisms.}} These approaches rely heavily on implicit alignment derived from large-scale pre-training rather than explicit transfer mechanisms. This dependency restricts rapid few-shot adaptation when encountering robots with novel kinematic structures.

To address these challenges, we propose MOTIF for efficient few-shot cross-embodiment transfer.  
Specifically, in Stage I, MOTIF encodes heterogeneous actions into unified action motifs via vector quantization (VQ)~\cite{VQ-VAE}. We employ a progress-aware alignment loss to enforce temporal consistency and an embodiment adversarial loss to bridge representational gaps.
In Stage II, we develop a lightweight multimodal motif predictor to infer appropriate action motifs conditioned on real-time observations and language instructions. 
In Stage III, we retrieve unified motifs and incorporate them into a vanilla flow-matching policy~\cite{flow-matching} to guide the action generation. The policy decodes predicted spatiotemporal motifs into actions by fusing them with embodiment-specific multimodal inputs, supporting efficient few-shot adaptation to new embodiments. 
Validated on heterogeneous robots via an interleaved task setting, MOTIF outperforms strong baselines by 6.5\% in simulation and 43.7\% in real-world few-shot scenarios, confirming the efficiency of action motif guidance.
Our contributions are summarized as follows:
\begin{itemize}
    \item We propose MOTIF, a hierarchical framework that achieves efficient few-shot cross-embodiment transfer by decoupling embodiment-agnostic spatiotemporal \textit{\textbf{action motifs}} from robot-specific execution.
    \item We introduce a unified motif learning mechanism incorporating progress-aware alignment and embodiment adversarial losses, coupled with a flow-matching policy to ground abstract motifs into precise actions.
    \item Extensive experiments demonstrate that MOTIF achieves state-of-the-art performance, surpassing strong baselines by \textbf{6.5\%} in simulation and \textbf{43.7\%} in real-world few-shot transfer.
\end{itemize}

\section{Related Work}
\subsection{Vision-Language-Action} 
Vision-language-action models (VLAs)~\cite{rt-2,Openvla,RDT-1B,VLA_survey} integrate computer vision and natural language processing into robotic control, aiming to map multimodal observations and text instructions directly to executable actions. This approach utilizes the rich semantic understanding of pre-trained foundation models to achieve open-vocabulary generalization in physical environments.
For example, RT-2~\cite{rt-2} discretizes robotic actions into text tokens, formulating control as a sequence modeling task to incorporate semantic knowledge from internet-scale pre-training. To mitigate precision loss in discrete tokenization, $\pi_0$~\cite{pi_0} integrates flow matching directly into VLMs to enable high-frequency continuous control. More recently, $\pi_{0.5}$~\cite{pi_0_5} incorporates explicit reasoning into the policy by generating intermediate tokens prior to action generation. This chain-of-thought process decomposes complex tasks into sub-goals, improving performance on long-horizon multi-step manipulation.
Despite these advancements, conventional VLAs focus on learning unified knowledge across embodiments but lack explicit cross-embodiment generalization designs. This limitation significantly increases the difficulty of transferring policies to downstream tasks on new embodiments.

\begin{figure*}[t]
  \begin{center}
    \centerline{\includegraphics[width=2.0\columnwidth]{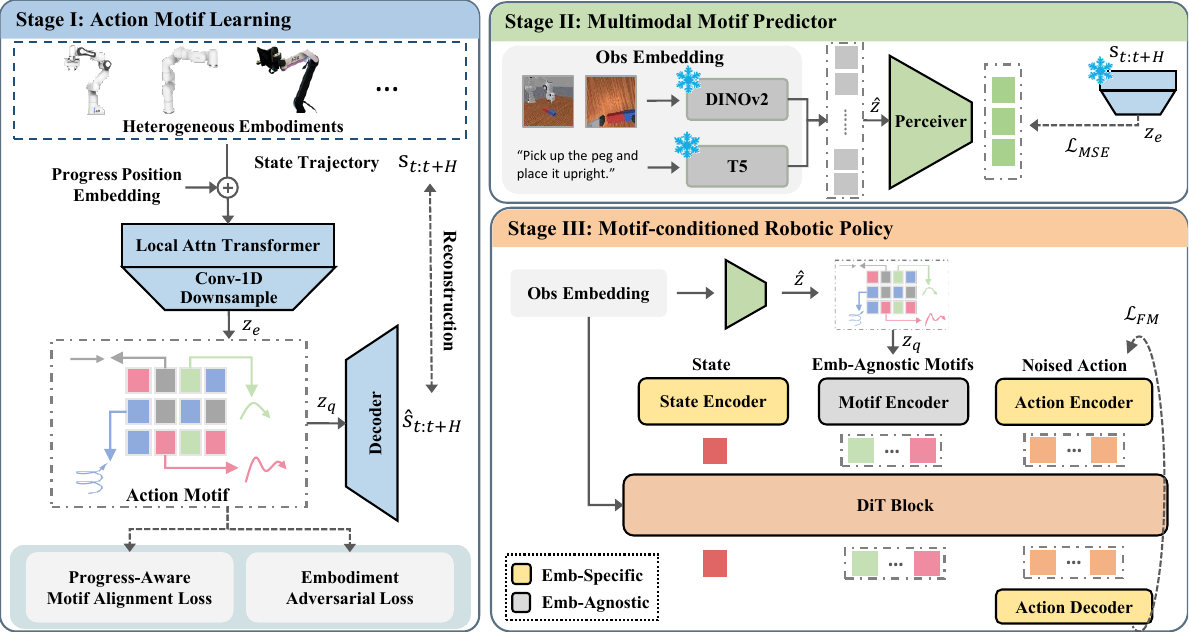}}
    \caption{\textbf{Overview of the MOTIF framework.} 
    \textbf{(Left) Stage I:} We learn unified action motifs from heterogeneous robot data using VQ-VAE augmented with Progress-Aware Alignment and Embodiment Adversarial objectives to ensure cross-embodiment consistency.
    \textbf{(Top Right) Stage II:} A multimodal predictor infers these motifs from vision and language inputs using frozen foundation encoders.
    \textbf{(Bottom Right) Stage III:} Inferred motifs serve as structural guidance for a flow-matching policy, enabling a Diffusion Transformer (DiT) to generate embodiment-specific actions via few-shot transfer.}
    \label{fig:framework} 
    \vspace{-8mm}
  \end{center}
\end{figure*}

\subsection{Cross-Embodiment Learning}
Cross-embodiment learning~\cite{CrossFormer,CrossSkill,UNIACT} aims to synthesize a unified policy capable of controlling diverse robots by learning a unified representation space to bridge kinematic discrepancies. Scaling this paradigm, RT-X~\cite{RT-X} aggregates datasets across diverse robotic platforms to train generalist policies, demonstrating that co-training with heterogeneous data improves robustness over single-robot baselines. To overcome architectural constraints, Heterogeneous Pretrained Transformers (HPT)~\cite{HPT} introduces a modular architecture designed to process variable-size proprioceptive and action inputs, allowing a single shared trunk to control robots with varying joint configurations without explicit alignment. Extending these principles to humanoid robotics, GR00T N1~\cite{GR00T-N1} establishes a foundation model tailored specifically for generalizable control and diverse locomotion tasks. However, these large-scale generalist policies typically require massive joint training and lack efficient mechanisms for few-shot adaptation to unseen embodiments, highlighting the need for the decoupled transfer approach proposed in this work.

Parallel research mitigates kinematic heterogeneity by learning unified latent representations that abstract away low-level execution details. Approaches like UniVLA~\cite{Univla} and GO-1~\cite{G0-1} construct task-centric or unified latent action spaces by mapping continuous visual observations to quantized representations, aligning cross-embodiment behaviors primarily based on visual state transitions. Conversely, focusing on the control domain, VQ-BeT~\cite{VQ-BET} and QueST~\cite{Quest} employ vector quantization on action data to derive discrete behavioral tokens, facilitating the modeling of multi-modal distributions. To further bridge perception and execution, XR-1~\cite{XR-1} introduces unified vision-motion codes that jointly learn from visual state transitions and action trajectories, enforcing alignment between visual dynamics and physical motion within a shared manifold. However, these approaches often rely on implicit alignment that remains entangled with source-domain kinematics or focus on broad standardization. They lack the explicit decoupling of spatiotemporal patterns from execution details, which is critical for data-efficient few-shot transfer to novel embodiments.

\section{Preliminaries} 
\subsection{Problem Formulation} 
\paragraph{Basic Setting.} We define multi-task cross-embodiment robot manipulation as learning a policy $\pi_\theta$ that maps the current language instruction $l$, observation image $o_t$ and state $s_t$ to a future action sequence executable by a specific embodiment:
\begin{equation}
    \label{eq:policy}
    \pi_\theta(l,o_t,s_t) \longrightarrow a_{t:t+H_a}\in\mathcal{A}_{e_i},
\end{equation}
where $a_{t:t+H_a}=\{a_t,a_{t+1},\dots,a_{t+H_a-1}\}$ denotes an action chunk of horizon $H_a$. In our setting, the set of embodiments $\mathcal{E}=\{e_1,e_2,\dots,e_N\}$ consists of robots with heterogeneous kinematic structures. For any embodiment $e_i \in \mathcal{E}$, its action $a_t$ belongs to an embodiment-specific action space $\mathcal{A}_{e_i}$, defined as the set of all feasible control signals for embodiment $e_i$. 
\paragraph{Cross-Embodiment Transfer.} Training data comprises extensive expert data $\mathcal{D}_{src}$ from source robots $\mathcal{E}_{src}$ and a few demonstrations $\mathcal{D}_{tgt}$ from the target robot $e_{tgt}$.
We aim to use these data to learn embodiment-agnostic action motifs and adapt policies to $\mathcal{A}_{e_{\text{tgt}}}$ under a few-shot setting.

\subsection{Flow Matching for Action Generation}
\label{sec:prelim-fm}
Flow Matching~\cite{flow-matching} has been widely adopted for action sequence generation in recent works~\cite{pi_0,VITA}.
Given the language instruction $l$, current observation $o_t$, and state $s_t$, the policy $\pi_\theta$ is instantiated as a conditional time-dependent velocity field $v_\theta(x \mid l, o_t, s_t)$ defined over a continuous time horizon $\tau \in [0,1]$.
The generative process is governed by the following ordinary differential equation:
\begin{equation}
d x_\tau / d \tau = v_\theta(x_\tau \mid l, o_t, s_t),
\end{equation}
which transports samples from Gaussian noise $x_0 \sim \mathcal{N}(0, I)$ to a future action chunk $x_1 = a_{t:t+H_a}$.

During training, we construct a linear interpolation path between $x_0$ and the expert action sequence $x_1$,
\begin{equation}
x_\tau = (1 - \tau)x_0 + \tau x_1,
\end{equation}
and supervise the network to match the corresponding ground-truth velocity field along this path.
Specifically, the ground-truth velocity is given by $x_1 - x_0$, and the flow-matching objective is defined as:
\begin{equation}
\mathcal{L}_{\text{FM}} = 
\mathbb{E}_{\tau, x_0, x_1}
\big[
\| v_\theta(x_\tau \mid l, o_t, s_t) - (x_1 - x_0) \|_2^2
\big].
\end{equation}

During inference, action sequences are generated by solving the learned ordinary differential equation from $\tau\!\!=\!\!0$ to $\tau\!\!=\!\!1$:
\begin{equation}
x_1 = x_0 + \int_{0}^{1} v_\theta(x_\tau \mid l, o_t, s_t)\, d\tau.
\end{equation}

\section{Method} 
In this section, we introduce MOTIF, a three-stage framework for few-shot cross-embodiment transfer, as illustrated in \cref{fig:framework}.
In Stage I, MOTIF learns embodiment-agnostic action motifs from short-horizon proprioceptive state segments across heterogeneous robots.
In Stage II, we train a lightweight multimodal motif predictor to infer motifs from visual observations and language instructions.
In Stage III, the predicted motifs condition a flow-matching policy as abstract spatiotemporal priors, guiding embodiment-specific action generation on target robots.
\begin{definition}[Action Motifs]
\label{def:delta-ordered}
We define action motifs as statistically significant trajectory subsequences that represent pure spatiotemporal patterns, independent of task semantics or robot embodiment.
\end{definition}

\subsection{Stage I: Action Motif Learning} 
\label{stage1}
In Stage I, our method learns unified embodiment-agnostic action motifs that capture reusable spatiotemporal patterns across heterogeneous robots.
We encode short-horizon proprioceptive state transitions into discrete latent representations, removing embodiment-specific kinematics.

\paragraph{Kinematic Trajectory Canonicalization.} 
To facilitate the learning of action motifs via vector quantization, preliminary kinematic alignment is a prerequisite.
To this end, we adopt fixed-window state trajectory segments as a kinematic-level action representation.
Given a segment $\{s_t, s_{t+1}, \dots, s_{t+H_s}\}$, the sequence describes the end-effector motion executed within the window.
We define each state $s_t$ as the absolute end-effector pose rather than joint configurations, since joint spaces exhibit strong embodiment-specific heterogeneity.
To mitigate these biases, we translate and rotate each end-effector trajectory into a canonical frame anchored at the initial end-effector pose, and apply scale normalization based on the robot workspace.
The resulting motion segment is denoted as:
\begin{equation}
    x = \mathcal{T}(s_{t:t+H_s}) \in \mathbb{R}^{H_s \times d_s},
\end{equation}
where $H_s$ denotes the trajectory horizon and $d_s$ represents the dimension of the state. This segment $x$ serves as the input to the latent action motif learning module.

\paragraph{Latent Action Motif Learning.} 
We learn discrete action motifs from motion segments of length $H_s$ using a vector-quantized autoencoder~\cite{VQ-VAE}.
Given a segment $x$, we incorporate a progress-aware positional encoding derived from the normalized timestamp within the demonstration into the motion features, enabling the model to distinguish motifs at different execution stages.

\begin{figure}[t]
  \begin{center}
    \centerline{\includegraphics[width=0.85\columnwidth]{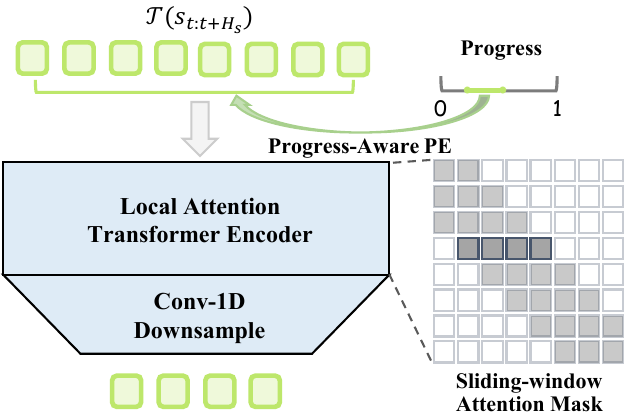}}
    \caption{\textbf{Architecture of the Latent Action Motif Learning Module.} 
    The encoder integrates progress-aware positional encodings (PE) and employs a local-attention Transformer with a sliding-window mask to capture local dynamics, followed by strided 1D convolution for temporal downsampling.}
    \label{fig:stage1_encoder}
  \end{center}
  \vskip -0.3in
\end{figure}

The encoder $E_{\phi}$, detailed in \cref{fig:stage1_encoder}, maps the input $x$ to a temporally downsampled token sequence of length $M$:
\begin{equation}
    z_e = E_{\phi}(x) = \{z_1, z_2, \dots, z_M\}, \quad z_m \in \mathbb{R}^{d_e}.
\end{equation}
To capture fine-grained local dynamics while maintaining computational efficiency, $E_{\phi}$ employs a local-attention Transformer with a restricted receptive field.
Specifically, we apply a sliding-window attention mask where each token at timestep $t$ attends exclusively to its symmetric local neighborhood $[t-k, t+k]$, to capture short-term kinematic dependencies.
Subsequently, a strided 1D convolutional layer compresses the temporal resolution from $H_s$ to $M$, yielding the compact latent representation sequence $z_e$. We discretize each token using a trainable codebook $\mathcal{C} = \{c_k\}_{k=1}^{K}$ with nearest-neighbor vector quantization:
\begin{equation}
    k_m = \operatorname*{argmin}_{j \in \{1,\dots,K\}} \| z_m - c_j \|_2, 
    \quad
    z_{q,m} = c_{k_m}.
\end{equation}
The decoder $\hat{x} = D_{\psi}(z_q)$ mirrors the encoder and reconstructs the motion sequence via temporal upsampling, ensuring that the discrete action motifs retain sufficient kinematic structure for accurate reconstruction.

We train the VQ-VAE by minimizing the standard objective:
\begin{equation}
\mathcal{L}_{\text{vq}} =
\| x - \hat{x} \|_2^2
+ \| \mathrm{sg}(z_e) - z_q \|_2^2
+ \beta \| z_e - \mathrm{sg}(z_q) \|_2^2 ,
\end{equation}
where $\mathrm{sg}(\cdot)$ denotes the stop-gradient operator, and $\beta$ is a hyperparameter that weights the commitment term.

\begin{algorithm}[tb]
\caption{Training and Inference Pipeline of MOTIF}
\label{alg:motif}
\begin{algorithmic}[1]
\REQUIRE Dataset $\mathcal{D}$, Pre-trained Encoders $f_{\text{img}}, f_{\text{lang}}$
\ENSURE Learned Policies $E_{\phi}, D_{\psi}, \mathcal{C}, R, v_{\theta}$

\STATE \textbf{Stage I: Action Motif Learning}
\FOR{sampled batch $x$ from $\mathcal{D}$}
    \STATE Encode: $z_e \leftarrow E_{\phi}(x)$
    \STATE Quantize: $z_q \leftarrow \text{VQ}(z_e; \mathcal{C})$  
    \STATE Reconstruct: $\hat{x} \leftarrow D_{\psi}(z_q)$      
    \STATE Update $\phi, \psi, \mathcal{C}, \omega$ to minimize $\mathcal{L}_{1}$
\ENDFOR

\STATE \textbf{Stage II: Multimodal Motif Predictor}
\STATE Freeze Stage I modules ($E_{\phi}, \mathcal{C}$)
\FOR{sampled batch $(o_t, l, x)$ from $\mathcal{D}$}
    \STATE Extract features: $h \leftarrow [f_{\text{img}}(o_t), f_{\text{lang}}(l)]$
    \STATE Predict tokens: $\hat{z} \leftarrow R_{\xi}(h)$
    \STATE Compute target: $z_m \leftarrow E_{\phi}(x)$
    \STATE Update $\xi$ to minimize $\mathcal{L}_{2}$
\ENDFOR

\STATE \textbf{Stage III: Motif-conditioned Robotic Policy}
\STATE Freeze Stage II predictor $R_{\xi}$
\FOR{sampled batch $(l, o_t, s_t, x)$ from $\mathcal{D}$}
    \STATE \textit{// 1. Infer Motif Prior}
    \STATE $\hat{z} \leftarrow R_{\xi}(f_{\text{img}}(o_t), f_{\text{lang}}(l))$
    \STATE $\tilde{z}_q \leftarrow \text{VQ}(\hat{z}; \mathcal{C})$
    
    \STATE \textit{// 2. Prepare Flow Matching}
    \STATE Sample $\tau \sim \mathcal{U}(0,1), \; x_1 \sim \mathcal{N}(0, I)$
    \STATE Construct input:   
    \STATE \quad $q_{\text{in}} \leftarrow \operatorname{Concat}(f_s(s_t), f_k(\tilde{z}_q), f_a(x_\tau))$ 
    \STATE Update $v_{\theta}$ to minimize $\mathcal{L}_{3}$ 
\ENDFOR

\STATE \textbf{Inference:}
\STATE Given $(l, o_t, s_t)$:
\STATE 1. Infer Motif:
\STATE \quad $\hat{z} \leftarrow R(f_{\text{img}}(o_t), f_{\text{lang}}(l)), \;\; \tilde{z}_q \leftarrow \text{VQ}(\hat{z}; \mathcal{C})$
\STATE 2. Generate Action:
\STATE \quad $x_1 = x_0 + \int_{0}^{1} v_\theta(x_\tau \mid s_t, o_t, l, \tilde{z}_q)\, d\tau$
\end{algorithmic}
\end{algorithm}

\paragraph{Progress-aware Motif Alignment Loss.} 
To ensure cross-embodiment kinematic alignment and temporal consistency of action motifs, we introduce a progress-aware objective that explicitly aligns motion segments of the same task phase.
We compute the normalized segment-level embedding $\hat{e}_i$ by averaging the token sequence $z_e$.
Within a batch, we prioritize aligning segments that share the same language instruction $l$ and occur at similar execution stages $p$. This is formalized by a progress-weighted similarity coefficient:
\begin{equation}
    w_{ij} = \mathbb{I}[l_i = l_j] \exp \left( - (|p_i - p_j| / \sigma)^2 \right),
\end{equation}
where $\sigma$ controls the temporal tolerance.
We then minimize the soft-weighted InfoNCE loss~\cite{infoNCE}:
\begin{equation}
\mathcal{L}_{\text{nce}}
= -\frac{1}{|\mathcal{A}|}\sum_{i\in\mathcal{A}}
\log
\frac{
\sum_{j\neq i} w_{ij}\,
\exp(\hat{e}_i^\top \hat{e}_j / \gamma)
}{
\sum_{k\neq i}
\exp(\hat{e}_i^\top \hat{e}_k / \gamma)
},
\end{equation}
where $\gamma$ is the temperature. This objective promotes embodiment-agnostic semantics by strictly aligning task-consistent and temporally synchronized segments.

\paragraph{Embodiment Adversarial Loss.}
To better model action motifs from embodiment-specific motion segments, we employ adversarial training via a gradient reversal layer (GRL)~\cite{GRL1,GRL2}.
We introduce an embodiment discriminator $D_{\omega}$ to identify robot identity $y$ from the latent tokens $\{z_m\}_{m=1}^{M}$.
The discriminator is trained to minimize the following objective:
\begin{equation}
    \mathcal{L}_{\text{adv}} = - \frac{1}{M} \sum_{m=1}^{M} \log D_{\omega}(y \mid z_m).
\end{equation}
During backpropagation, the GRL inverts the gradient flow to the encoder, effectively forcing it to generate embodiment-invariant representations that confuse the discriminator.

\paragraph{Overall training objective.}
The joint optimization objective for Stage I is:
\begin{equation}
\mathcal{L}_{1} =
\mathcal{L}_{\text{vq}}
+ \lambda_{\text{nce}} \mathcal{L}_{\text{nce}}
- \lambda_{\text{adv}} \mathcal{L}_{\text{adv}}.
\end{equation}
In our implementation, we set $\beta = 0.25$, $\lambda_{\text{nce}} = 0.1$ and $\lambda_{\text{adv}} = 0.1$ to balance the motif alignment and adversarial objectives.
This objective balances reconstruction fidelity, motif alignment, and embodiment invariance for learning unified action motifs.

\begin{figure*}[t]
  \begin{center}
    \centerline{\includegraphics[width=2.0\columnwidth]{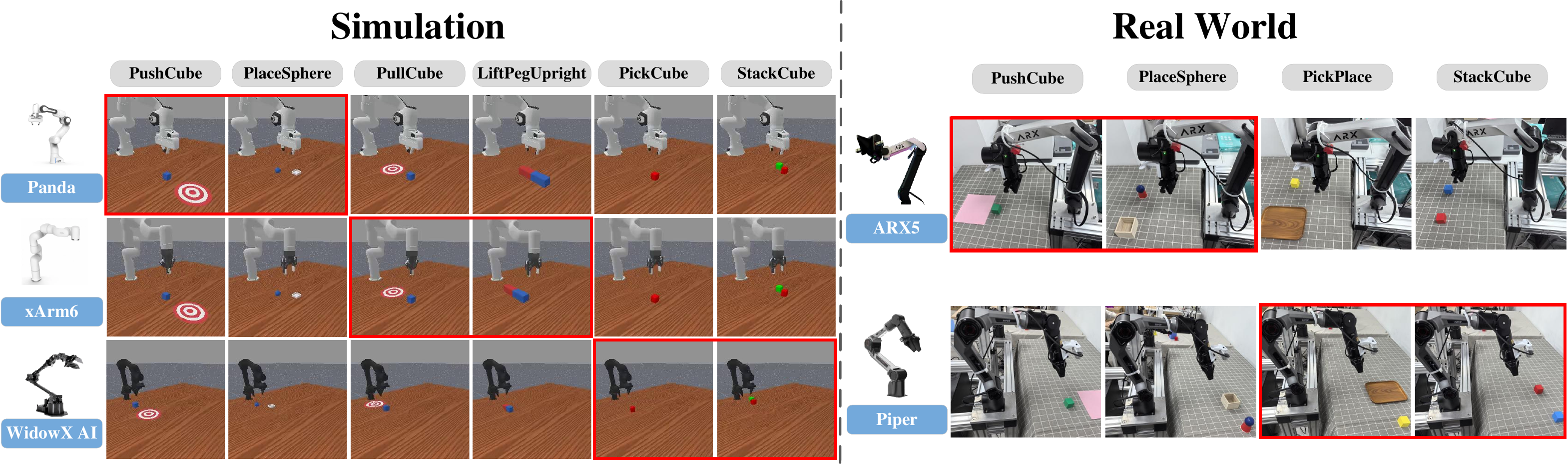}}
    \caption{\textbf{Overview of Simulation and Real-World Environments.} 
  We evaluate MOTIF across heterogeneous embodiments in both simulated (Left) and physical (Right) settings. 
  The experiments follow an interleaved task allocation protocol, where red bounding boxes (\protect\redbox) denote the target (few-shot) embodiment-task pairs used to assess cross-embodiment transfer capability. 
  The remaining pairs serve as the source domain with full supervision.}
    \label{fig:env_overview}
    \vspace{-10pt}
  \end{center}
\end{figure*}

\subsection{Stage II: Multimodal Motif Predictor} 
\label{stage2}
In Stage II, we construct a lightweight multimodal motif predictor designed to infer appropriate action motifs based on real-time observations and language instructions.
This policy bridges the gap between observation and latent action space, enabling test-time motif inference without access to the future state trajectories.

Given a current observation $o_t$ and a language instruction $l$, we first utilize frozen pre-trained encoders to extract rich semantic features.
Specifically, we employ a DINOv2~\cite{dinov2} vision encoder $f_{\text{img}}$ and a T5~\cite{t5} language encoder $f_{\text{lang}}$.
These multimodal features are fused and compressed via a perceiver~\cite{perceiver} module $R_{\xi}$ into a fixed-length sequence of $M$ predicted motif tokens, denoted as $\hat{z} = \{\hat{z}_1, \dots, \hat{z}_M\}$:
\begin{equation}
    \hat{z} = R_{\xi}\big( [f_{\text{img}}(o_t), f_{\text{lang}}(l)] \big).
\end{equation}
During training, we regress the predicted tokens $\hat{z}$ to the ground-truth encoder representations $z_e=\{z_1,\dots,z_M\}$ derived from the frozen Stage I encoder using an MSE loss:
\begin{equation}
    \mathcal{L}_{2} = \frac{1}{M}\sum_{m=1}^{M} \| \hat{z}_m - \operatorname{sg}(z_m) \|_2^2.
\end{equation}
This alignment mechanism equips Stage III with robust motif priors inferred solely from vision and language, serving as structural guidance for action generation.
Specifically, the predicted continuous tokens $\hat{z}$ are quantized via the codebook $\mathcal{C}$ to obtain discrete motif embeddings which guide the subsequent robotic policy.

\subsection{Stage III: Motif-conditioned Robotic Policy} 
\label{stage3}
In Stage III, we integrate unified embodiment-agnostic action motifs with embodiment-specific control to generate executable action sequences.
We condition a flow-matching policy with predicted action motifs as a structural prior, guiding action generation toward intended motions under the kinematic constraints of the target embodiment.

\paragraph{Retrieving Discrete Motifs.}
We obtain the discrete action motifs $\tilde{z}_q$ by querying the unified codebook $\mathcal{C}$ using the continuous tokens predicted by the frozen Stage II predictor. This nearest-neighbor quantization ensures that the structural guidance strictly aligns with the embodiment-agnostic latent space established in Stage I.

\paragraph{Motif-conditioned Action Generation.}
We parameterize the policy using a standard flow-matching diffusion transformer (DiT)~\cite{diffusion_transformer}. We employ embodiment-specific encoders for the proprioceptive state $s_t$ and the noised action chunk $x_\tau$, while the discrete motif sequence $\tilde{z}_q$ is mapped by a shared encoder $f_k$.
We construct the DiT input tokens $q_{\text{in}}$ by concatenating these embeddings:
\begin{equation}
    q_{\text{in}} = \mathrm{Concat}\big(f_s(s_t),\; f_k(\tilde{z}_q),\; f_a(x_\tau)\big).
\end{equation}
Further, we derive the cross-modal conditioning context $c$ from the frozen vision and language encoders:
\begin{equation}
    c = \mathrm{Concat}\big(f_{\text{img}}(o_t),\; f_{\text{lang}}(l)\big).
\end{equation}
Inside the DiT blocks, the input tokens $q_{\text{in}}$ serve as the query to attend to the context $c$.
The policy is trained to minimize the conditional flow-matching loss:
\begin{equation}
    \mathcal{L}_{3} = \mathbb{E}_{\tau, x_0, x_1} \big[ \| v_\theta(x_\tau | l, o_t, s_t, \tilde{z}_q) - (x_1 \!-\! x_0) \|_2^2 \big].
\end{equation}

\begin{table*}[t]
\caption{\textbf{Multi-task Cross-Embodiment Transfer Results (Simulation).} We report success rates (\%) across varying supervision levels ($K \in \{1, 3, 5, 10, 50\}$). 
\textbf{Transfer} measures the average success rate on target (few-shot) pairs, indicating cross-embodiment transfer. 
\textbf{Global} measures the overall performance across all 18 pairs. 
MOTIF significantly outperforms baselines in the data-scarce transfer regimes. 
The symbol * denotes that methods are pretrained on large-scale robot datasets.}
\label{tab:main_results}
\begin{center}
\begin{small}
\resizebox{\textwidth}{!}{
\begin{tabular}{l|c|cc|cc|cc|cc|cc}
\toprule
\multirow{2}{*}{\textbf{Method}} & \multirow{2}{*}{\textbf{Params (B)}} & \multicolumn{2}{c|}{\textbf{1-Shot}} & \multicolumn{2}{c|}{\textbf{3-Shot}} & \multicolumn{2}{c|}{\textbf{5-Shot}} & \multicolumn{2}{c|}{\textbf{10-Shot}} & \multicolumn{2}{c}{\textbf{50-Shot}} \\
 & & \textbf{Transfer} & \textbf{Global} & \textbf{Transfer} & \textbf{Global} & \textbf{Transfer} & \textbf{Global} & \textbf{Transfer} & \textbf{Global} & \textbf{Transfer} & \textbf{Global} \\
\midrule
Diffusion Policy & 0.22 & 15.67 & 27.00 & 27.00 & 31.33 & 29.33 & 30.44 & 38.00 & 36.11 & 46.00 & 38.00 \\
HPT$^{*}$        & 0.06 & 10.00 & 21.11 & 15.00 & 23.44 & 17.33 & 26.33 & 26.33 & 29.67 & 44.33 & 35.11 \\
$\pi_0^{*}$      & 3.30 & \underline{33.33} & \underline{45.78} & \underline{38.33} & 48.22 & \underline{45.67} & \underline{52.67} & \underline{56.33} & \underline{59.56} & \underline{67.67} & \underline{62.89} \\
GR00T N1$^{*}$   & 2.00 & 21.67 & 43.67 & 33.00 & \underline{50.67} & 35.00 & 50.44 & 44.67 & 50.78 & 57.67 & 57.44 \\
\midrule
\rowcolor{gray!15} \textbf{MOTIF (Ours)} & \textbf{0.31} & \textbf{36.00} & \textbf{55.78} & \textbf{48.33} & \textbf{55.33} & \textbf{54.33} & \textbf{60.44} & \textbf{60.33} & \textbf{60.44} & \textbf{75.00} & \textbf{66.00} \\
\bottomrule
\end{tabular}
}
\end{small}
\end{center}
\vskip -0.1in
\end{table*}

\section{Experiments}

In this section, we evaluate MOTIF in both simulation and real-world settings on multi-task few-shot cross-embodiment manipulation, as visualized in \cref{fig:env_overview}. We test whether embodiment-agnostic action motifs mitigate cross-embodiment misalignment by transferring task-relevant spatiotemporal structure across heterogeneous robots.

We adopt an interleaved task setting, where each embodiment has full demonstrations for a subset of tasks and only a few demonstrations for the remaining tasks, enforcing cross-embodiment transfer under limited target supervision.
We aim to answer the following questions:

\begin{enumerate}
    \item \textbf{Few-shot transfer:} Can MOTIF improve success rates with limited demonstrations (e.g., 1/3/5-shot) compared to end-to-end baselines without action motifs?

    \item \textbf{Ablations on action motifs:} How much do action motifs contribute to performance? Specifically, how do (i) kinematic trajectory canonicalization, (ii) progress-aware alignment, and (iii) embodiment adversarial objectives affect motif quality and transferability?

    \item \textbf{Real-world deployment:} Can MOTIF transfer effectively in real-world manipulation and generate physically executable actions under practical conditions?
\end{enumerate}

\subsection{Simulation Setups and Baselines}

\textbf{Simulation Setups.} 
We evaluate on the ManiSkill~\cite{maniskill3} benchmark due to its support for diverse robot embodiments, enabling the assessment of transfer under kinematic heterogeneity. Our experimental setup utilizes three distinct robots: Franka Panda, xArm6, and WidowX AI. These robots perform six diverse manipulation tasks, including \texttt{PushCube}, \texttt{PlaceSphere}, \texttt{PullCube}, \texttt{LiftPegUpright}, \texttt{PickCube}, and \texttt{StackCube}. Detailed task descriptions and specific success criteria are provided in Appendix~\ref{sec:app_sim_tasks}. We collect 50 expert demonstrations for each embodiment-task pair, providing a standardized basis for the interleaved task evaluation.

\textbf{Interleaved Task Setting.}
To evaluate few-shot cross-embodiment transfer, we adopt an \textit{interleaved task mask protocol}.
As detailed in \cref{tab:task_allocation}, we partition tasks for each embodiment into Source tasks with \textbf{Full} data and Target tasks with \textbf{Few} data. \textbf{Few} refers to the specific number of demonstrations $K$ provided for transfer, where $K$ takes values from the set $\{1, 3, 5, 10, 50\}$. For these few-shot cases, we strictly select the first $K$ demonstrations from the full sequence of 50 episodes rather than random sampling. This distribution ensures that every task serves as a full data source on one robot while remaining a few-shot target on another. This configuration prevents simple in-domain memorization and provides a rigorous benchmark to verify the effectiveness of cross-embodiment transfer.

\begin{table}[t]
\caption{\textbf{Interleaved Task Allocation.} We adopt an interleaved protocol where tasks are partitioned into Source (\full, 50 demos) and Target (\few, $K$ demos). \textbf{(a)} Simulation setup with 6 tasks across 3 robots. \textbf{(b)} Real-world setup with 4 tasks across 2 robots. The \few\ entries denote the target embodiment-task pairs used for few-shot training and evaluation.}
\label{tab:task_allocation}
\begin{center}

\footnotesize 
\setlength{\tabcolsep}{3.5pt} 

\textbf{(a) Simulation Environment} \vspace{2pt} \\
\begin{tabular}{lcccccc}
\toprule
\textbf{Robot} & \mcell{Push\\Cube} & \mcell{Place\\Sphere} & \mcell{Pull\\Cube} & \mcell{LiftPeg\\Upright} & \mcell{Pick\\Cube} & \mcell{Stack\\Cube} \\
\midrule
Panda     & \few & \few & \full & \full & \full & \full \\
xArm6     & \full & \full & \few & \few & \full & \full \\
WidowX AI & \full & \full & \full & \full & \few & \few \\
\bottomrule
\end{tabular}

\vspace{3mm} 

\textbf{(b) Real World Environment} \vspace{2pt} \\
\begin{tabular}{lcccc}
\toprule
\textbf{Robot} & \mcell{Push\\Cube} & \mcell{Place\\Sphere} & \mcell{Pick\\Cube} & \mcell{Stack\\Cube} \\
\midrule
ARX5  & \few & \few & \full & \full \\
Piper & \full & \full & \few & \few \\
\bottomrule
\end{tabular}
\end{center}
\vskip -0.2in
\end{table}

\textbf{Baselines.}
We compare MOTIF with several representative methods, including Diffusion Policy~\cite{Diffusion-Policy}, $\pi_0$~\cite{pi_0}, HPT~\cite{HPT}, and GR00T-N1~\cite{GR00T-N1}. These baselines cover both end-to-end action generation and shared-private embodiment conditioning paradigms. All are trained and evaluated under the same interleaved protocol and few-shot conditions.

\subsection{Simulation Experiments}

\textbf{Evaluation Metrics.}
\label{Evaluation_Metrics.}
We evaluate policy performance using the \textbf{Success Rate (SR)}, averaged over 50 rollouts for each embodiment-task pair. Let $\mathcal{S}$ denote the set of all embodiment-task pairs, and $\mathcal{S}_{\text{Few}} \subset \mathcal{S}$ denote the subset of target pairs (marked as ``\few'' in \cref{tab:task_allocation}) restricted to few-shot supervision.
Let $R(e, \tau)$ be the success rate for embodiment $e$ on task $\tau$. We report two aggregated metrics:
\begin{equation}
\small
    \text{Global} \!=\! \frac{1}{|\mathcal{S}|} \!\!\!\sum_{(e, \tau) \in \mathcal{S}}\!\!\!\!\! R(e, \tau),
    \text{Transfer} \!=\! \frac{1}{|\mathcal{S}_{\text{Few}}|} \!\!\!\!\sum_{(e, \tau) \in \mathcal{S}_{\text{Few}}}\!\!\!\!\!\!\!\! R(e, \tau).
\end{equation}

The \textbf{Global} metric reflects the overall mastery of both source and target skills, whereas \textbf{Transfer} measures the capability to adapt to target embodiment-task combinations under data-scarce conditions.

\textbf{Quantitative Results.}
As shown in Table~\ref{tab:main_results}, MOTIF outperforms baselines, particularly in data-scarce regimes.
In the challenging 1-shot setting, both scratch-trained methods (e.g., Diffusion Policy) and pre-trained models (e.g., HPT, GR00T N1) struggle to generalize, achieving success rates below 22\%. While $\pi_0$ performs better (33.33\%), MOTIF still achieves the highest performance of \textbf{36.00\%}.
This advantage becomes more pronounced at 5-shot, where MOTIF rapidly improves to \textbf{54.33\%}, demonstrating a much steeper learning curve than $\pi_0$ (45.67\%) and GR00T N1 (35.00\%).
Even with full supervision ($K=50$), MOTIF maintains distinct superiority (\textbf{75.00\%}).
These results confirm that retrieving unified action motifs provides a structural prior for rapid adaptation to target kinematics.

\begin{table}[t]
\caption{\textbf{Ablation on Motif Guidance.} We compare the Transfer success rate (\%) with and without incorporating retrieved motifs in Stage III. The removal of motif guidance leads to consistent performance drops across all supervision levels.}
\label{tab:ablation_motif}
\begin{center}
\begin{small}
\resizebox{\columnwidth}{!}{
\begin{tabular}{l|ccccc}
\toprule
\textbf{Setting} & \textbf{1-Shot} & \textbf{3-Shot} & \textbf{5-Shot} & \textbf{10-Shot} & \textbf{50-Shot} \\
\midrule
w/o Motif Guidance & 30.67 & 43.67 & 47.33 & 58.00 & 71.67 \\
\rowcolor{gray!15} \textbf{MOTIF (Full)} & \textbf{36.00} & \textbf{48.33} & \textbf{54.33} & \textbf{60.33} & \textbf{75.00} \\
\bottomrule
\end{tabular}
}
\end{small}
\end{center}
\vskip -0.1in
\end{table}

\subsection{Ablation Study.}
To validate the effectiveness of key components, we conduct ablation studies focusing on the Transfer success rate.

\textbf{Effectiveness of Motif Guidance.} 
We first investigate the impact of incorporating retrieved motifs during Stage III. As shown in \cref{tab:ablation_motif}, removing motif guidance consistently degrades performance across all data regimes. 
Notably, in the low-data settings (1-shot and 3-shot), the performance drops by 5.33\% and 4.66\%, respectively. 
This indicates that without the explicit structural prior provided by the retrieved motifs, the policy struggles to efficiently ground high-level intents to target kinematics, leading to slower transfer.

\textbf{Impact of Action Motif Learning Designs.} 
We investigate the essential components in Stage I for constructing the embodiment-agnostic motifs. 
Constructing the motif space without Kinematic Canonicalization causes the sharpest decline of 10.33\%, confirming that unifying motion representations is a prerequisite for effective retrieval across embodiments. 
Additionally, our training objectives are crucial for refinement: removing the Progress-aware Alignment Loss ($\mathcal{L}_{\text{nce}}$) and Embodiment Adversarial Loss ($\mathcal{L}_{\text{adv}}$) leads to drops of 4.66\% and 2.66\%, respectively. 
These results demonstrate that $\mathcal{L}_{\text{nce}}$ ensures temporal consistency while $\mathcal{L}_{\text{adv}}$ helps eliminate embodiment-specific features, collectively ensuring the robustness of the learned motifs.

\begin{table}[t]
\caption{\textbf{Ablation on Model Components (5-Shot).} We analyze the impact of Kinematic Trajectory Canonicalization (KTC), Progress-aware Alignment Loss ($\mathcal{L}_{\text{nce}}$), and Embodiment Adversarial Loss ($\mathcal{L}_{\text{adv}}$). Results indicate that each component is indispensable for effective cross-embodiment transfer.}
\label{tab:ablation_components}
\begin{center}
\begin{small}
\resizebox{1.0\columnwidth}{!}{
\begin{tabular}{l|c|c}
\toprule
\textbf{Variant} & \textbf{Transfer SR (\%)} & \textbf{$\Delta$} \\
\midrule
\rowcolor{gray!15} \textbf{MOTIF (Full)} & \textbf{54.33} & - \\
\midrule
w/o Kinematic Canonicalization & 44.00 & -10.33 \\
w/o Alignment Loss ($\mathcal{L}_{\text{nce}}$) & 49.67 & -4.66 \\
w/o Adversarial Loss ($\mathcal{L}_{\text{adv}}$) & 51.67 & -2.66 \\
\bottomrule
\end{tabular}
}
\end{small}
\end{center}
\vskip -0.1in
\end{table}

\subsection{Real-World Experiments}
\textbf{Experiment Setups.} 
To validate the effectiveness of MOTIF in real-world environments, we employ two distinct robotic arms: the \textbf{ARX5} and the \textbf{Piper} arm.
We constructed a real-world dataset covering four manipulation tasks: \textit{PickPlace}, \textit{PushCube}, \textit{StackCube}, and \textit{PlaceSphere}.
For each embodiment-task pair, we collect 50 expert demonstrations.
Consistent with our simulation benchmark, we adopt the same Interleaved Task Allocation protocol as detailed in Table~\ref{tab:task_allocation}(b) to evaluate cross-embodiment transfer.
The policy performance is reported using the same metrics defined in \ref{Evaluation_Metrics.}, averaged over 20 rollouts for each setting.

\begin{table}[t]
\caption{\textbf{Real-World Cross-Embodiment Evaluation (5-Shot).} Success rates (\%) over 20 rollouts per task, following Target settings in Table~1(b). Transfer and Global averages are reported.}
\label{tab:real_world}
\begin{center}
\begin{small}
\renewcommand{\arraystretch}{1.15} 
\resizebox{\columnwidth}{!}{
\begin{tabular}{l|c|cccc|cc}
\toprule
\multirow{2}{*}{\textbf{Method}} & \multirow{2}{*}{\textbf{Robot}} & \multicolumn{4}{c|}{\textbf{Tasks}} & \multicolumn{2}{c}{\textbf{Metrics}} \\
 & & \textbf{\makecell{Push\\Cube}} & \textbf{\makecell{Place\\Sphere}} & \textbf{\makecell{Pick\\Place}} & \textbf{\makecell{Stack\\Cube}} & \textbf{Transfer} & \textbf{Global} \\
\midrule
\multirow{2}{*}{Diffusion Policy} & ARX5 & 5.0 & 30.0 & 40.0 & 50.0 & \multirow{2}{*}{23.75} & \multirow{2}{*}{36.88} \\
 & Piper & 55.0 & 55.0 & 55.0 & 5.0 & & \\
\midrule
\multirow{2}{*}{GR00T N1$^{*}$} & ARX5 & 10.0 & 20.0 & 40.0 & 50.0 & \multirow{2}{*}{21.25} & \multirow{2}{*}{33.13} \\
 & Piper & 50.0 & 40.0 & 40.0 & 15.0 & & \\
\midrule

\rowcolor{gray!15} \textbf{MOTIF} & ARX5 & 60.0 & 90.0 & 65.0 & 100.0 & & \\ 
\rowcolor{gray!15} \textbf{(Ours)} & Piper & 70.0 & 90.0 & 70.0 & 50.0 & \multirow{-2}{*}{\textbf{67.50}} & \multirow{-2}{*}{\textbf{74.38}} \\ 
\bottomrule
\end{tabular}
}
\end{small}
\end{center}
\vskip -0.2in
\end{table}

\textbf{Quantitative Results.} \cref{tab:real_world} reports the 5-shot transfer performance on physical ARX5 and Piper arms. 
The gap between MOTIF and baselines is even more pronounced in the real world than in simulation. 
MOTIF achieves a Transfer Avg. of \textbf{67.50\%}, substantially outperforming Diffusion Policy (23.75\%) and GR00T N1 (21.25\%). 
Unlike baselines that degrade under hardware noise, MOTIF effectively grounds retrieved motifs into executable actions, demonstrating real-world robustness.

\section{Conclusion}
We introduced MOTIF, a framework enabling efficient few-shot cross-embodiment transfer by decoupling embodiment-agnostic action motifs from robot-specific kinematics. By leveraging progress-aware vector quantization and a motif-conditioned flow-matching policy, MOTIF effectively aligns heterogeneous action spaces with minimal target data. Extensive experiments in simulation and the real world demonstrate that our approach significantly outperforms strong baselines, validating the importance of explicit structural priors for scalable generalist robotic learning.

\bibliography{example_paper}
\bibliographystyle{icml2025}

\newpage
\appendix
\onecolumn

\section{Implementation and Training Details}
\label{sec:app_implementation}

\subsection{Hyperparameter Settings}
\label{sec:app_hyperparams}
We detail the model architectures and specific hyperparameters for Action Motif Learning (Stage I), the Multimodal Motif Predictor (Stage II), and the Motif-conditioned Robotic Policy (Stage III) in \cref{tab:model_config}. The optimization settings are largely consistent across stages, utilizing the AdamW optimizer with a cosine learning rate schedule. All experiments for the three stages are conducted on a single NVIDIA RTX 4090 GPU. We summarize the common and stage-specific training hyperparameters in \cref{tab:optim_params}.

\begin{table*}[htbp]
    \caption{\textbf{Model Configurations for MOTIF Framework.}}
    \label{tab:model_config}
    \centering
    \begin{small}
    \renewcommand{\arraystretch}{1.25}
    
    \begin{tabular}{c|c|c}
    \hline
    \textbf{Stage I: Action Motif Learning} & \textbf{Stage II: Motif Predictor} & \textbf{Stage III: Robotic Policy} \\
    \hline
    
    \begin{tabular}[t]{@{}l l@{}}
        \multicolumn{2}{@{}l}{\textit{\textbf{Architecture (VQ-VAE)}}} \\
        Window Size ($H_s$) & 32 \\
        Motif Num ($M$) & 16 \\
        Model Dim ($d_{model}$) & 256 \\
        Latent Dim ($d_e$) & 256 \\
        Codebook Size ($K$) & 128 \\
        Enc/Dec Layers & 4 / 4 \\
        Enc/Dec Heads & 8 / 8 \\
        Dropout & 0.1 \\
        Conv Layers & 2 \\
        Kernel Sizes & [5,3] \\
        Strides & [2,1] \\
        Local Neighborhood ($k$) & 8 \\
        \multicolumn{2}{@{}l}{\textit{\textbf{Loss Coefficients}}} \\
        Commitment ($\beta$) & 0.25 \\
        Alignment ($\lambda_{\text{nce}}$) & 0.1 \\
        Adversarial ($\lambda_{\text{adv}}$) & 0.1 \\
    \end{tabular}
    
    & 
    
    \begin{tabular}[t]{@{}l l@{}}
        \multicolumn{2}{@{}l}{\textit{\textbf{Encoders (Frozen)}}} \\
        Vision & DINOv2 \\
        Language & T5-Base \\
        \multicolumn{2}{@{}l}{\textit{\textbf{Perceiver Resampler}}} \\
        Model Dim & 512 \\
        Depth & 6 \\
        Heads & 8 \\
        Dim Head & 64 \\
        Latent Num ($M$) & 16 \\
    \end{tabular}
    
    & 
    
    \begin{tabular}[t]{@{}l l@{}}
        \multicolumn{2}{@{}l}{\textit{\textbf{Architecture (DiT)}}} \\
        Action Horizon ($H_a$) & 16 \\
        Hidden Size & 512 \\
        Num Layers & 16 \\
        Num Heads & 8 \\
        Norm Type & AdaNorm \\
        Dropout & 0.2 \\
        \multicolumn{2}{@{}l}{\textit{\textbf{Flow Matching}}} \\
        Noise Beta ($\alpha, \beta$) & (1.5, 1.0) \\
        Noise Scale ($s$) & 0.999 \\
        Inf. Timesteps & 4 \\
        Buckets & 1000 \\
    \end{tabular} \\
    \hline
    \end{tabular}
    \end{small}
\end{table*}

\begin{table}[h]
    \caption{\textbf{Optimization Hyperparameters.}}
    \label{tab:optim_params}
    \centering
    \begin{small}
    \renewcommand{\arraystretch}{1.2}
    \begin{tabular}{l|ccc}
    \toprule
    \textbf{Hyperparameter} & \textbf{Stage I} & \textbf{Stage II} & \textbf{Stage III} \\
    \midrule
    \multicolumn{4}{l}{\textit{\textbf{Common Settings}}} \\
    Optimizer & \multicolumn{3}{c}{AdamW} \\
    Peak Learning Rate & \multicolumn{3}{c}{1e-4} \\
    Weight Decay & \multicolumn{3}{c}{0.01} \\
    Warmup Ratio & \multicolumn{3}{c}{0.05} \\
    Gradient Clip Norm & \multicolumn{3}{c}{1.0} \\
    \midrule
    \multicolumn{4}{l}{\textit{\textbf{Stage-Specific Settings}}} \\
    Batch Size & 128 & 128 & 64 \\
    Training Epochs & 20 & 30 & 60 \\
    \bottomrule
    \end{tabular}
    \end{small}
\end{table}

\subsection{Baseline Implementation Details}
\label{sec:app_baselines}

We provide additional details of baselines used in our experiments.
Overall, our baselines cover (i) generative behavior cloning policies for continuous control, and (ii) cross-embodiment architectures that explicitly support heterogeneous proprioception and action spaces.
All baselines are trained and evaluated under the same interleaved task protocol and few-shot budgets.

\textbf{Diffusion Policy (DP)~\cite{Diffusion-Policy}.} 
Diffusion Policy is a generative behavior cloning method that models the conditional distribution of action trajectories via a denoising diffusion process. 
In our implementation, we utilize the codebase from the LeRobot~\cite{lerobot} platform. 
We configure the policy with an observation history of 2 frames ($n_{\text{obs\_steps}}=2$), a prediction horizon of 16 ($H=16$), and an execution chunk size of 8 ($n_{\text{action\_steps}}=8$). 
The model is trained for 200,000 steps on a single NVIDIA RTX 5090 GPU with a batch size of 64, utilizing random cropping with a 90\% ratio for visual data augmentation.

\textbf{\boldmath$\pi_0$\unboldmath~\cite{pi_0}.} 
$\pi_0$ is a flow-matching-based vision-language-action model that integrates a VLM backbone to enable high-frequency continuous control. 
Following the official fine-tuning protocol, we perform full fine-tuning on the Base $\pi_0$ model using a single NVIDIA RTX Pro 6000 GPU. 
The training is conducted for 100,000 steps with a batch size of 32, utilizing a learning rate schedule with 3,000 warmup steps, a peak learning rate of $2 \times 10^{-5}$, and a decay learning rate of $1.5 \times 10^{-6}$.

\textbf{Heterogeneous Pretrained Transformers (HPT)~\cite{HPT}.} 
HPT is a modular policy architecture designed for cross-embodiment learning. 
It employs a ``stem-trunk-head'' structure, where embodiment-specific \textit{stems} encode heterogeneous proprioceptive states into a unified latent space, a shared Transformer \textit{trunk} processes these latents alongside visual tokens, and embodiment-specific \textit{heads} decode the features into actionable control signals.
In our implementation, we select the \texttt{HPT-Large} version of the model.
To support simultaneous multi-robot training, we instantiate distinct stems and heads for each embodiment. 
We perform full fine-tuning on all modules (including the shared trunk, as we observed superior performance) for 1,000 epochs. 
The training is conducted on a single NVIDIA RTX 4090 GPU with a batch size of 768 and a learning rate of $1.0 \times 10^{-5}$.

\textbf{GR00T-N1~\cite{GR00T-N1}.} 
GR00T-N1 is a generalist vision-language-action policy designed for scalable cross-embodiment control. 
Architecturally, it employs a shared Diffusion Transformer (DiT) backbone to model the flow-based action distribution, coupled with lightweight embodiment-specific \textit{projectors} (comprising state encoders, action encoders, and action decoders) to bridge heterogeneous kinematic spaces. 
This design is structurally analogous to our Stage~III policy, making it the most direct baseline to validate the efficacy of our proposed \emph{action motif} guidance. 
In our experiments, we adhere to the official fine-tuning protocol, training the model for 100,000 steps on a single NVIDIA L40 GPU. 
The training utilizes a batch size of 32 and a learning rate of $1.0 \times 10^{-4}$.

\textbf{Fair comparison.}
For all baselines, we use the same demonstration datasets, few-shot target budgets, and evaluation protocols.
All methods operate on the same observation modalities, including proprioceptive states, third-person and wrist-view images, and language instructions (with the exception of Diffusion Policy, which does not support language conditioning).

\section{Experimental Setups}
\label{sec:app_setup}

\subsection{Simulation Task Description}
\label{sec:app_sim_tasks}

We provide detailed descriptions and success criteria for the six manipulation tasks used in ManiSkill~\cite{maniskill,maniskill3} simulation experiments.

\textbf{PushCube:} The goal is to push a cube into a designated target region. Success is achieved when the cube is fully contained within the target area.

\textbf{PlaceSphere:} The goal is to pick a sphere and place it into a container. Success is achieved when the sphere is stably positioned inside the container.

\textbf{PullCube:} The goal is to pull a cube into a designated target region. Success is achieved when the cube is fully contained within the target area.

\textbf{LiftPegUpright:} The goal is to reorient a peg to stand upright on the table. Success is achieved when the peg remains stable in a vertical position.

\textbf{PickCube:} The goal is to grasp a cube and lift it off the table. Success is achieved when the cube is lifted more than 2.5 cm above the surface.

\textbf{StackCube:} The goal is to stack one cube on top of another. Success is achieved when the top cube is stable and the gripper is released.

\subsection{Real-World Setup and Data Collection}
\label{sec:app_real_world}
\begin{figure*}[t]
  \centering 
  \includegraphics[width=1.0\textwidth]{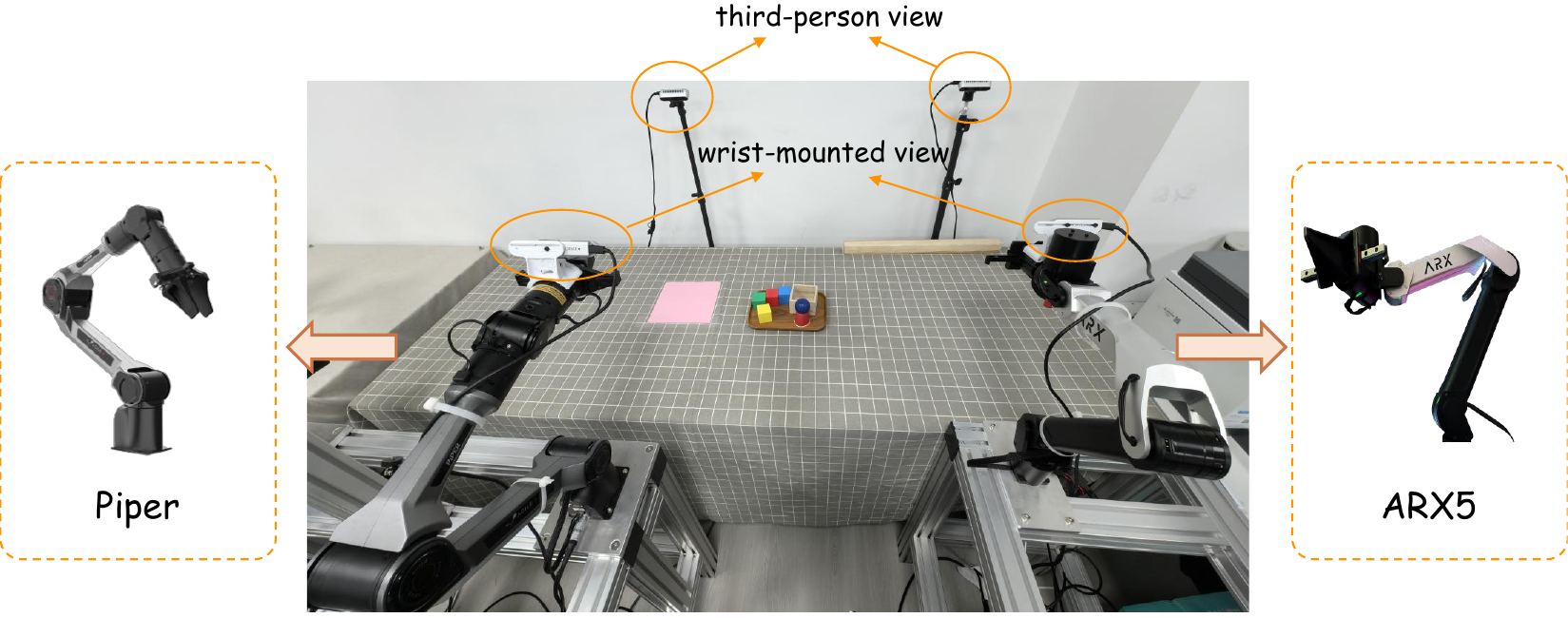}
  \caption{\textbf{Hardware Setup.} We evaluate MOTIF on two heterogeneous single-arm robots: Piper (left) and ARX5 (right).}
  \label{fig:real_setup}
\end{figure*}

\textbf{Hardware Setup.} 
As illustrated in \cref{fig:real_setup}, our real-world evaluation employs two distinct single-arm robots: the \textbf{ARX5} and the \textbf{Piper}. 
To ensure consistent visual perception across embodiments, each robot is equipped with two Intel RealSense D435i cameras, capturing both third-person and wrist-mounted views.

\textbf{Data Collection.}
We collect expert demonstrations using a \textbf{leader-follower teleoperation system}.
Visual observations are captured at a resolution of 640$\times$480, and the data collection frequency is set to 15Hz.
The collected dataset adheres to the interleaved task protocol detailed in \cref{tab:task_allocation} of the main paper.
Specifically, corresponding to \cref{tab:task_allocation}(b), we collect 50 demonstrations for Source tasks (marked as ``Full'') and the specified number of shots for Target tasks (marked as ``Few'') to rigorously evaluate few-shot transfer capabilities.

\textbf{Real-World Task Description.}
We evaluate MOTIF on four real-world manipulation tasks involving diverse interactions:
\textit{PushCube}: Push the green cube into the pink area.
\textit{PlaceSphere}: Pick up the ball and place into the box.
\textit{PickPlace}: Pick up the cube and place on the plate.
\textit{StackCube}: Stack the red cube on the blue cube.

\section{Additional Experimental Results}
\label{sec:app_additional_results}

\subsection{Detailed Simulation Results}
\label{sec:detailed_sim_results}

We present the detailed per-task success rates for all methods across all supervision levels ($K \in \{1, 3, 5, 10, 50\}$).
In the following tables, \textbf{cells highlighted in yellow} denote the target embodiment-task pairs (corresponding to the ``Few'' split in the interleaved protocol), where the model is adapted using only $K$ demonstrations. Conversely, uncolored cells represent source tasks with full supervision (50 demos).

To provide a comprehensive analysis, we report four aggregated metrics:
\begin{itemize}
    \item \textbf{Global:} The average success rate across all six tasks for a specific robot row.
    \item \textbf{Transfer:} The average success rate calculated exclusively on the target tasks (yellow cells) for a specific robot, measuring few-shot adaptation performance on that embodiment.
    \item \textbf{Cross-Emb. Global:} The macro-average of the \textbf{Global} metric across all three heterogeneous robots (Panda, xArm6, WidowX) at a specific shot level.
    \item \textbf{Cross-Emb. Transfer:} The macro-average of the \textbf{Transfer} metric across all three robots. This serves as the primary indicator for evaluating the method's capability in cross-embodiment few-shot generalization.
\end{itemize}

\cref{tab:dp_results}, \cref{tab:hpt_results}, \cref{tab:groot_n1_results}, \cref{tab:pi0_results}, and \cref{tab:motif_results} provide the complete performance for Diffusion Policy~\cite{Diffusion-Policy}, HPT~\cite{HPT}, GR00T N1~\cite{GR00T-N1}, $\pi_0$~\cite{pi_0}, and MOTIF, respectively.

\begin{center}
  \captionof{table}{Diffusion Policy success rates (\%) on six tasks across embodiments under few-shot settings.}
  \label{tab:dp_results}
  \centering
  \small
  \renewcommand{\arraystretch}{1.25}
  \resizebox{\textwidth}{!}{%
    \begin{tabular}{m{2cm}|l|lcccccc|c|c|c|c}
      \Xhline{1.2pt}
      \multicolumn{1}{c|}{\textbf{\makecell{Models}}} &
      \textbf{\makecell{Shots}} &
      \textbf{\makecell{Robot}} &
      \makecell{PushCube} &
      \makecell{PlaceSphere} &
      \makecell{PullCube} &
      \makecell{LiftPegUpright} &
      \makecell{PickCube} &
      \makecell{StackCube} &
      \makecell{Global} &
      \makecell{Cross-Emb.\\Global} &
      \makecell{Transfer} &
      \makecell{Cross-Emb.\\Transfer} \\
      \hline

      \multirow{15}{*}{\textbf{Diffusion Policy}}
      & \multirow{3}{*}{1}  & Panda    & \cellcolor[HTML]{FEFCD6}56.00\% & \cellcolor[HTML]{FEFCD6}6.00\% & 86.00\% & 16.00\% &  0.00\% &  0.00\% & 27.33\%
      & \multirow{3}{*}{27.00\%} & 31.00\% & \multirow{3}{*}{15.67\%} \\
      &                         & xArm6    & 92.00\% &  2.00\% & \cellcolor[HTML]{FEFCD6}28.00\% & \cellcolor[HTML]{FEFCD6}0.00\% & 12.00\% &  8.00\% & 23.67\%
      &                            & 14.00\% &                            \\
      &                         & WidowX AI & 80.00\% &  2.00\% & 38.00\% & 56.00\% & \cellcolor[HTML]{FEFCD6}4.00\% & \cellcolor[HTML]{FEFCD6}0.00\% & 30.00\%
      &                            &  2.00\% &                            \\
      \cline{2-13}

      & \multirow{3}{*}{3}  & Panda    & \cellcolor[HTML]{FEFCD6}38.00\% & \cellcolor[HTML]{FEFCD6}0.00\% & 92.00\% & 44.00\% & 20.00\% &  4.00\% & 33.00\%
      & \multirow{3}{*}{31.33\%} & 19.00\% & \multirow{3}{*}{27.00\%} \\
      &                         & xArm6    & 84.00\% &  2.00\% & \cellcolor[HTML]{FEFCD6}44.00\% & \cellcolor[HTML]{FEFCD6}0.00\% &  2.00\% &  0.00\% & 22.00\%
      &                            & 22.00\% &                            \\
      &                         & WidowX AI & 74.00\% &  4.00\% &  6.00\% & 70.00\% & \cellcolor[HTML]{FEFCD6}80.00\% & \cellcolor[HTML]{FEFCD6}0.00\% & 39.00\%
      &                            & 40.00\% &                            \\
      \cline{2-13}

      & \multirow{3}{*}{5}  & Panda    & \cellcolor[HTML]{FEFCD6}34.00\% & \cellcolor[HTML]{FEFCD6}8.00\% & 76.00\% & 36.00\% & 20.00\% &  0.00\% & 29.00\%
      & \multirow{3}{*}{30.44\%} & 21.00\% & \multirow{3}{*}{29.33\%} \\
      &                         & xArm6    & 78.00\% &  2.00\% & \cellcolor[HTML]{FEFCD6}72.00\% & \cellcolor[HTML]{FEFCD6}0.00\% &  4.00\% &  0.00\% & 26.00\%
      &                            & 36.00\% &                            \\
      &                         & WidowX AI & 62.00\% &  6.00\% & 20.00\% & 68.00\% & \cellcolor[HTML]{FEFCD6}60.00\% & \cellcolor[HTML]{FEFCD6}2.00\% & 36.33\%
      &                            & 31.00\% &                            \\
      \cline{2-13}

      & \multirow{3}{*}{10} & Panda    & \cellcolor[HTML]{FEFCD6}86.00\% & \cellcolor[HTML]{FEFCD6}2.00\% & 84.00\% & 40.00\% & 14.00\% &  0.00\% & 37.67\%
      & \multirow{3}{*}{36.11\%} & 44.00\% & \multirow{3}{*}{38.00\%} \\
      &                         & xArm6    & 86.00\% & 14.00\% & \cellcolor[HTML]{FEFCD6}72.00\% & \cellcolor[HTML]{FEFCD6}4.00\% &  0.00\% & 12.00\% & 31.33\%
      &                            & 38.00\% &                            \\
      &                         & WidowX AI & 78.00\% &  0.00\% & 28.00\% & 66.00\% & \cellcolor[HTML]{FEFCD6}60.00\% & \cellcolor[HTML]{FEFCD6}4.00\% & 39.33\%
      &                            & 32.00\% &                            \\
      \cline{2-13}

      & \multirow{3}{*}{50} & Panda    & \cellcolor[HTML]{FEFCD6}94.00\% & \cellcolor[HTML]{FEFCD6}6.00\% & 82.00\% & 34.00\% & 16.00\% &  0.00\% & 38.67\%
      & \multirow{3}{*}{38.00\%} & 50.00\% & \multirow{3}{*}{46.00\%} \\
      &                         & xArm6    & 84.00\% &  0.00\% & \cellcolor[HTML]{FEFCD6}94.00\% & \cellcolor[HTML]{FEFCD6}0.00\% &  8.00\% &  2.00\% & 31.33\%
      &                            & 47.00\% &                            \\
      &                         & WidowX AI & 56.00\% & 10.00\% & 40.00\% & 76.00\% & \cellcolor[HTML]{FEFCD6}80.00\% & \cellcolor[HTML]{FEFCD6}2.00\% & 44.00\%
      &                            & 41.00\% &                            \\
      \Xhline{1.2pt}
    \end{tabular}%
  }
  \vskip -0.1in
\end{center}

\begin{center}
  \captionof{table}{HPT success rates (\%) on six tasks across embodiments under few-shot settings.}
  \label{tab:hpt_results}
  \centering
  \small
  \renewcommand{\arraystretch}{1.25}
  \resizebox{\textwidth}{!}{%
    \begin{tabular}{m{2cm}|l|lcccccc|c|c|c|c}
      \Xhline{1.2pt}
      \multicolumn{1}{c|}{\textbf{\makecell{Models}}} &
      \textbf{\makecell{Shots}} &
      \textbf{\makecell{Robot}} &
      \makecell{PushCube} &
      \makecell{PlaceSphere} &
      \makecell{PullCube} &
      \makecell{LiftPegUpright} &
      \makecell{PickCube} &
      \makecell{StackCube} &
      \makecell{Global} &
      \makecell{Cross-Emb.\\Global} &
      \makecell{Transfer} &
      \makecell{Cross-Emb.\\Transfer} \\
      \hline

      \multirow{15}{*}{\textbf{HPT}}
      & \multirow{3}{*}{1}  & Panda    & \cellcolor[HTML]{FEFCD6}16.00\% & \cellcolor[HTML]{FEFCD6}0.00\% & 52.00\% &  0.00\% & 48.00\% &  0.00\% & 19.33\%
      & \multirow{3}{*}{21.11\%} &  8.00\% & \multirow{3}{*}{10.00\%} \\
      &                         & xArm6    & 44.00\% & 18.00\% & \cellcolor[HTML]{FEFCD6}0.00\% & \cellcolor[HTML]{FEFCD6}4.00\% & 64.00\% &  0.00\% & 21.67\%
      &                            &  2.00\% &                            \\
      &                         & WidowX AI & 12.00\% & 14.00\% & 68.00\% &  0.00\% & \cellcolor[HTML]{FEFCD6}40.00\% & \cellcolor[HTML]{FEFCD6}0.00\% & 22.33\%
      &                            & 20.00\% &                            \\
      \cline{2-13}

      & \multirow{3}{*}{3}  & Panda    & \cellcolor[HTML]{FEFCD6}22.00\% & \cellcolor[HTML]{FEFCD6}4.00\% & 58.00\% &  4.00\% & 70.00\% &  0.00\% & 26.33\%
      & \multirow{3}{*}{23.44\%} & 13.00\% & \multirow{3}{*}{15.00\%} \\
      &                         & xArm6    & 18.00\% & 26.00\% & \cellcolor[HTML]{FEFCD6}0.00\% & \cellcolor[HTML]{FEFCD6}2.00\% & 76.00\% &  8.00\% & 21.67\%
      &                            &  1.00\% &                            \\
      &                         & WidowX AI &  6.00\% & 24.00\% & 42.00\% &  0.00\% & \cellcolor[HTML]{FEFCD6}62.00\% & \cellcolor[HTML]{FEFCD6}0.00\% & 22.33\%
      &                            & 31.00\% &                            \\
      \cline{2-13}

      & \multirow{3}{*}{5}  & Panda    & \cellcolor[HTML]{FEFCD6}30.00\% & \cellcolor[HTML]{FEFCD6}4.00\% & 64.00\% & 10.00\% & 76.00\% &  2.00\% & 31.00\%
      & \multirow{3}{*}{26.33\%} & 17.00\% & \multirow{3}{*}{17.33\%} \\
      &                         & xArm6    & 30.00\% & 24.00\% & \cellcolor[HTML]{FEFCD6}6.00\% & \cellcolor[HTML]{FEFCD6}2.00\% & 70.00\% &  2.00\% & 22.33\%
      &                            &  4.00\% &                            \\
      &                         & WidowX AI & 26.00\% & 20.00\% & 42.00\% &  4.00\% & \cellcolor[HTML]{FEFCD6}60.00\% & \cellcolor[HTML]{FEFCD6}2.00\% & 25.67\%
      &                            & 31.00\% &                            \\
      \cline{2-13}

      & \multirow{3}{*}{10} & Panda    & \cellcolor[HTML]{FEFCD6}36.00\% & \cellcolor[HTML]{FEFCD6}6.00\% & 70.00\% &  2.00\% & 70.00\% &  2.00\% & 31.00\%
      & \multirow{3}{*}{29.67\%} & 21.00\% & \multirow{3}{*}{26.33\%} \\
      &                         & xArm6    & 38.00\% & 28.00\% & \cellcolor[HTML]{FEFCD6}22.00\% & \cellcolor[HTML]{FEFCD6}12.00\% & 72.00\% &  2.00\% & 29.00\%
      &                            & 17.00\% &                            \\
      &                         & WidowX AI & 26.00\% & 14.00\% & 52.00\% &  0.00\% & \cellcolor[HTML]{FEFCD6}80.00\% & \cellcolor[HTML]{FEFCD6}2.00\% & 29.00\%
      &                            & 41.00\% &                            \\
      \cline{2-13}

      & \multirow{3}{*}{50} & Panda    & \cellcolor[HTML]{FEFCD6}70.00\% & \cellcolor[HTML]{FEFCD6}22.00\% & 54.00\% &  8.00\% & 56.00\% &  2.00\% & 35.33\%
      & \multirow{3}{*}{35.11\%} & 46.00\% & \multirow{3}{*}{44.33\%} \\
      &                         & xArm6    & 40.00\% & 28.00\% & \cellcolor[HTML]{FEFCD6}68.00\% & \cellcolor[HTML]{FEFCD6}12.00\% & 48.00\% &  0.00\% & 32.67\%
      &                            & 40.00\% &                            \\
      &                         & WidowX AI & 28.00\% & 12.00\% & 44.00\% & 46.00\% & \cellcolor[HTML]{FEFCD6}94.00\% & \cellcolor[HTML]{FEFCD6}0.00\% & 37.33\%
      &                            & 47.00\% &                            \\
      \Xhline{1.2pt}
    \end{tabular}%
  }
  \vskip -0.1in
\end{center}

\begin{center}
  \captionof{table}{GR00T N1 success rates (\%) on six tasks across embodiments under few-shot settings.}
  \label{tab:groot_n1_results}
  \centering
  \small
  \renewcommand{\arraystretch}{1.25}
  \resizebox{\textwidth}{!}{%
    \begin{tabular}{m{2cm}|l|lcccccc|c|c|c|c}
      \Xhline{1.2pt}
      \multicolumn{1}{c|}{\textbf{\makecell{Models}}} &
      \textbf{\makecell{Shots}} &
      \textbf{\makecell{Robot}} &
      \makecell{PushCube} &
      \makecell{PlaceSphere} &
      \makecell{PullCube} &
      \makecell{LiftPegUpright} &
      \makecell{PickCube} &
      \makecell{StackCube} &
      \makecell{Global} &
      \makecell{Cross-Emb.\\Global} &
      \makecell{Transfer} &
      \makecell{Cross-Emb.\\Transfer} \\
      \hline

      \multirow{15}{*}{\textbf{GR00T N1}}
      & \multirow{3}{*}{1}  & Panda    & \cellcolor[HTML]{FEFCD6}44.00\% & \cellcolor[HTML]{FEFCD6}0.00\% & 56.00\% & 30.00\% & 84.00\% & 22.00\% & 39.33\%
      & \multirow{3}{*}{43.67\%} & 22.00\% & \multirow{3}{*}{21.67\%} \\
      &                         & xArm6    & 76.00\% & 48.00\% & \cellcolor[HTML]{FEFCD6}24.00\% & \cellcolor[HTML]{FEFCD6}8.00\% & 98.00\% & 26.00\% & 46.67\%
      &                            & 16.00\% &                            \\
      &                         & WidowX AI & 72.00\% & 30.00\% & 52.00\% & 62.00\% & \cellcolor[HTML]{FEFCD6}54.00\% & \cellcolor[HTML]{FEFCD6}0.00\% & 45.00\%
      &                            & 27.00\% &                            \\
      \cline{2-13}

      & \multirow{3}{*}{3}  & Panda    & \cellcolor[HTML]{FEFCD6}50.00\% & \cellcolor[HTML]{FEFCD6}10.00\% & 54.00\% & 52.00\% & 84.00\% & 22.00\% & 45.33\%
      & \multirow{3}{*}{50.67\%} & 30.00\% & \multirow{3}{*}{33.00\%} \\
      &                         & xArm6    & 78.00\% & 60.00\% & \cellcolor[HTML]{FEFCD6}32.00\% & \cellcolor[HTML]{FEFCD6}28.00\% & 100.00\% & 26.00\% & 54.00\%
      &                            & 30.00\% &                            \\
      &                         & WidowX AI & 74.00\% & 30.00\% & 66.00\% & 68.00\% & \cellcolor[HTML]{FEFCD6}78.00\% & \cellcolor[HTML]{FEFCD6}0.00\% & 52.67\%
      &                            & 39.00\% &                            \\
      \cline{2-13}

      & \multirow{3}{*}{5}  & Panda    & \cellcolor[HTML]{FEFCD6}52.00\% & \cellcolor[HTML]{FEFCD6}4.00\% & 60.00\% & 42.00\% & 84.00\% & 22.00\% & 44.00\%
      & \multirow{3}{*}{50.44\%} & 28.00\% & \multirow{3}{*}{35.00\%} \\
      &                         & xArm6    & 78.00\% & 56.00\% & \cellcolor[HTML]{FEFCD6}34.00\% & \cellcolor[HTML]{FEFCD6}28.00\% & 100.00\% & 28.00\% & 54.00\%
      &                            & 31.00\% &                            \\
      &                         & WidowX AI & 76.00\% & 20.00\% & 72.00\% & 60.00\% & \cellcolor[HTML]{FEFCD6}90.00\% & \cellcolor[HTML]{FEFCD6}2.00\% & 53.33\%
      &                            & 46.00\% &                            \\
      \cline{2-13}

      & \multirow{3}{*}{10} & Panda    & \cellcolor[HTML]{FEFCD6}88.00\% & \cellcolor[HTML]{FEFCD6}16.00\% & 58.00\% & 32.00\% & 82.00\% & 24.00\% & 50.00\%
      & \multirow{3}{*}{50.78\%} & 52.00\% & \multirow{3}{*}{44.67\%} \\
      &                         & xArm6    & 80.00\% & 56.00\% & \cellcolor[HTML]{FEFCD6}52.00\% & \cellcolor[HTML]{FEFCD6}26.00\% & 100.00\% & 18.00\% & 55.33\%
      &                            & 39.00\% &                            \\
      &                         & WidowX AI & 68.00\% & 10.00\% & 56.00\% & 62.00\% & \cellcolor[HTML]{FEFCD6}82.00\% & \cellcolor[HTML]{FEFCD6}4.00\% & 47.00\%
      &                            & 43.00\% &                            \\
      \cline{2-13}

      & \multirow{3}{*}{50} & Panda    & \cellcolor[HTML]{FEFCD6}100.00\% & \cellcolor[HTML]{FEFCD6}32.00\% & 52.00\% & 40.00\% & 82.00\% & 26.00\% & 55.33\%
      & \multirow{3}{*}{57.44\%} & 66.00\% & \multirow{3}{*}{57.67\%} \\
      &                         & xArm6    & 90.00\% & 58.00\% & \cellcolor[HTML]{FEFCD6}98.00\% & \cellcolor[HTML]{FEFCD6}8.00\% & 100.00\% & 12.00\% & 61.00\%
      &                            & 53.00\% &                            \\
      &                         & WidowX AI & 66.00\% & 32.00\% & 64.00\% & 66.00\% & \cellcolor[HTML]{FEFCD6}96.00\% & \cellcolor[HTML]{FEFCD6}12.00\% & 56.00\%
      &                            & 54.00\% &                            \\
      \Xhline{1.2pt}
    \end{tabular}%
  }
  \vskip -0.1in
\end{center}

\vspace{50mm}

\begin{center}
  \captionof{table}{$\pi_0$ success rates (\%) on six tasks across embodiments under few-shot settings.}
  \label{tab:pi0_results}
  \centering
  \small
  \renewcommand{\arraystretch}{1.25}
  \resizebox{\textwidth}{!}{%
    \begin{tabular}{m{2cm}|l|lcccccc|c|c|c|c}
      \Xhline{1.2pt}
      \multicolumn{1}{c|}{\textbf{\makecell{Models}}} &
      \textbf{\makecell{Shots}} &
      \textbf{\makecell{Robot}} &
      \makecell{PushCube} &
      \makecell{PlaceSphere} &
      \makecell{PullCube} &
      \makecell{LiftPegUpright} &
      \makecell{PickCube} &
      \makecell{StackCube} &
      \makecell{Global} &
      \makecell{Cross-Emb.\\Global} &
      \makecell{Transfer} &
      \makecell{Cross-Emb.\\Transfer} \\
      \hline

      \multirow{15}{*}{\scalebox{1.15}{\textbf{\boldmath$\pi_0$}}}
      & \multirow{3}{*}{1}  & Panda    & \cellcolor[HTML]{FEFCD6}66.00\% & \cellcolor[HTML]{FEFCD6}30.00\% & 88.00\% & 20.00\% & 84.00\% & 34.00\% & 53.67\%
      & \multirow{3}{*}{45.78\%} & 48.00\% & \multirow{3}{*}{33.33\%} \\
      &                          & xArm6    & 84.00\% & 44.00\% & \cellcolor[HTML]{FEFCD6}32.00\% & \cellcolor[HTML]{FEFCD6}0.00\% & 94.00\% & 28.00\% & 47.00\%
      &                             & 16.00\% &                            \\
      &                          & WidowX AI & 48.00\% & 24.00\% & 40.00\% & 36.00\% & \cellcolor[HTML]{FEFCD6}68.00\% & \cellcolor[HTML]{FEFCD6}4.00\% & 36.67\%
      &                             & 36.00\% &                            \\
      \cline{2-13}

      & \multirow{3}{*}{3}  & Panda    & \cellcolor[HTML]{FEFCD6}48.00\% & \cellcolor[HTML]{FEFCD6}46.00\% & 80.00\% & 16.00\% & 80.00\% & 32.00\% & 50.33\%
      & \multirow{3}{*}{48.22\%} & 47.00\% & \multirow{3}{*}{38.33\%} \\
      &                          & xArm6    & 78.00\% & 54.00\% & \cellcolor[HTML]{FEFCD6}46.00\% & \cellcolor[HTML]{FEFCD6}10.00\% & 86.00\% & 34.00\% & 51.33\%
      &                             & 28.00\% &                            \\
      &                          & WidowX AI & 36.00\% & 38.00\% & 42.00\% & 62.00\% & \cellcolor[HTML]{FEFCD6}78.00\% & \cellcolor[HTML]{FEFCD6}2.00\% & 43.00\%
      &                             & 40.00\% &                            \\
      \cline{2-13}

      & \multirow{3}{*}{5}  & Panda    & \cellcolor[HTML]{FEFCD6}74.00\% & \cellcolor[HTML]{FEFCD6}56.00\% & 76.00\% & 18.00\% & 78.00\% & 34.00\% & 56.00\%
      & \multirow{3}{*}{52.67\%} & 65.00\% & \multirow{3}{*}{45.67\%} \\
      &                          & xArm6    & 80.00\% & 66.00\% & \cellcolor[HTML]{FEFCD6}66.00\% & \cellcolor[HTML]{FEFCD6}10.00\% & 80.00\% & 28.00\% & 55.00\%
      &                             & 38.00\% &                            \\
      &                          & WidowX AI & 62.00\% & 24.00\% & 58.00\% & 70.00\% & \cellcolor[HTML]{FEFCD6}68.00\% & \cellcolor[HTML]{FEFCD6}0.00\% & 47.00\%
      &                             & 34.00\% &                            \\
      \cline{2-13}

      & \multirow{3}{*}{10} & Panda    & \cellcolor[HTML]{FEFCD6}70.00\% & \cellcolor[HTML]{FEFCD6}42.00\% & 78.00\% & 50.00\% & 86.00\% & 42.00\% & 61.33\%
      & \multirow{3}{*}{59.56\%} & 56.00\% & \multirow{3}{*}{56.33\%} \\
      &                          & xArm6    & 80.00\% & 68.00\% & \cellcolor[HTML]{FEFCD6}84.00\% & \cellcolor[HTML]{FEFCD6}52.00\% & 84.00\% & 32.00\% & 66.67\%
      &                             & 68.00\% &                            \\
      &                          & WidowX AI & 62.00\% & 32.00\% & 50.00\% & 70.00\% & \cellcolor[HTML]{FEFCD6}88.00\% & \cellcolor[HTML]{FEFCD6}2.00\% & 50.67\%
      &                             & 45.00\% &                            \\
      \cline{2-13}

      & \multirow{3}{*}{50} & Panda    & \cellcolor[HTML]{FEFCD6}90.00\% & \cellcolor[HTML]{FEFCD6}82.00\% & 80.00\% & 48.00\% & 90.00\% & 48.00\% & 73.00\%
      & \multirow{3}{*}{62.89\%} & 86.00\% & \multirow{3}{*}{67.67\%} \\
      &                          & xArm6    & 76.00\% & 66.00\% & \cellcolor[HTML]{FEFCD6}78.00\% & \cellcolor[HTML]{FEFCD6}34.00\% & 84.00\% & 42.00\% & 63.33\%
      &                             & 56.00\% &                            \\
      &                          & WidowX AI & 50.00\% & 48.00\% & 40.00\% & 54.00\% & \cellcolor[HTML]{FEFCD6}98.00\% & \cellcolor[HTML]{FEFCD6}24.00\% & 52.33\%
      &                             & 61.00\% &                            \\
      \Xhline{1.2pt}
    \end{tabular}%
  }
  \vskip -0.1in
\end{center}

\begin{center}
  \captionof{table}{\textbf{MOTIF} success rates (\%) on six tasks across embodiments under few-shot settings (same metrics as Table~\ref{tab:dp_results}).}
  \label{tab:motif_results}
  \centering
  \small
  \renewcommand{\arraystretch}{1.25}
  \resizebox{\textwidth}{!}{%
    \begin{tabular}{m{2cm}|l|lcccccc|c|c|c|c}
      \Xhline{1.2pt}
      \multicolumn{1}{c|}{\textbf{\makecell{Models}}} &
      \textbf{\makecell{Shots}} &
      \textbf{\makecell{Robot}} &
      \makecell{PushCube} &
      \makecell{PlaceSphere} &
      \makecell{PullCube} &
      \makecell{LiftPegUpright} &
      \makecell{PickCube} &
      \makecell{StackCube} &
      \makecell{Global} &
      \makecell{Cross-Emb.\\Global} &
      \makecell{Transfer} &
      \makecell{Cross-Emb.\\Transfer} \\
      \hline

      \multirow{15}{*}{\textbf{MOTIF}}
      & \multirow{3}{*}{1}  & Panda    & \cellcolor[HTML]{FEFCD6}98.00\% & \cellcolor[HTML]{FEFCD6}14.00\% & 94.00\% & 18.00\% & 96.00\% & 60.00\% & 63.33\%
      & \multirow{3}{*}{55.78\%} & 56.00\% & \multirow{3}{*}{36.00\%} \\
      &                         & xArm6    & 80.00\% & 90.00\% & \cellcolor[HTML]{FEFCD6}70.00\% & \cellcolor[HTML]{FEFCD6}2.00\% & 100.00\% & 66.00\% & 68.00\%
      &                            & 36.00\% &                            \\
      &                         & WidowX AI & 56.00\% & 38.00\% & 50.00\% & 40.00\% & \cellcolor[HTML]{FEFCD6}30.00\% & \cellcolor[HTML]{FEFCD6}2.00\% & 36.00\%
      &                            & 16.00\% &                            \\
      \cline{2-13}

      & \multirow{3}{*}{3}  & Panda    & \cellcolor[HTML]{FEFCD6}94.00\% & \cellcolor[HTML]{FEFCD6}26.00\% & 96.00\% & 28.00\% & 72.00\% & 42.00\% & 59.67\%
      & \multirow{3}{*}{55.33\%} & 60.00\% & \multirow{3}{*}{48.33\%} \\
      &                         & xArm6    & 86.00\% & 90.00\% & \cellcolor[HTML]{FEFCD6}54.00\% & \cellcolor[HTML]{FEFCD6}16.00\% & 100.00\% & 76.00\% & 70.33\%
      &                            & 35.00\% &                            \\
      &                         & WidowX AI & 34.00\% & 42.00\% & 18.00\% & 22.00\% & \cellcolor[HTML]{FEFCD6}98.00\% & \cellcolor[HTML]{FEFCD6}2.00\% & 36.00\%
      &                            & 50.00\% &                            \\
      \cline{2-13}

      & \multirow{3}{*}{5}  & Panda    & \cellcolor[HTML]{FEFCD6}96.00\% & \cellcolor[HTML]{FEFCD6}44.00\% & 90.00\% & 28.00\% & 84.00\% & 70.00\% & 68.67\%
      & \multirow{3}{*}{60.44\%} & 70.00\% & \multirow{3}{*}{54.33\%} \\
      &                         & xArm6    & 84.00\% & 88.00\% & \cellcolor[HTML]{FEFCD6}68.00\% & \cellcolor[HTML]{FEFCD6}24.00\% & 94.00\% & 64.00\% & 70.33\%
      &                            & 46.00\% &                            \\
      &                         & WidowX AI & 60.00\% & 46.00\% & 32.00\% & 22.00\% & \cellcolor[HTML]{FEFCD6}88.00\% & \cellcolor[HTML]{FEFCD6}6.00\% & 42.33\%
      &                            & 47.00\% &                            \\
      \cline{2-13}

      & \multirow{3}{*}{10} & Panda    & \cellcolor[HTML]{FEFCD6}98.00\% & \cellcolor[HTML]{FEFCD6}50.00\% & 98.00\% & 28.00\% & 90.00\% & 52.00\% & 69.33\%
      & \multirow{3}{*}{60.44\%} & 74.00\% & \multirow{3}{*}{60.33\%} \\
      &                         & xArm6    & 84.00\% & 80.00\% & \cellcolor[HTML]{FEFCD6}72.00\% & \cellcolor[HTML]{FEFCD6}42.00\% & 94.00\% & 54.00\% & 71.00\%
      &                            & 57.00\% &                            \\
      &                         & WidowX AI & 64.00\% & 30.00\% &  8.00\% & 44.00\% & \cellcolor[HTML]{FEFCD6}90.00\% & \cellcolor[HTML]{FEFCD6}10.00\% & 41.00\%
      &                            & 50.00\% &                            \\
      \cline{2-13}

      & \multirow{3}{*}{50} & Panda    & \cellcolor[HTML]{FEFCD6}100.00\% & \cellcolor[HTML]{FEFCD6}72.00\% & 94.00\% & 48.00\% & 96.00\% & 56.00\% & 77.33\%
      & \multirow{3}{*}{66.67\%} & 85.00\% & \multirow{3}{*}{71.67\%} \\
      &                         & xArm6    & 76.00\% & 94.00\% & \cellcolor[HTML]{FEFCD6}96.00\% & \cellcolor[HTML]{FEFCD6}42.00\% & 94.00\% & 72.00\% & 79.00\%
      &                            & 69.00\% &                            \\
      &                         & WidowX AI & 72.00\% &  2.00\% & 60.00\% &  6.00\% & \cellcolor[HTML]{FEFCD6}100.00\% & \cellcolor[HTML]{FEFCD6}22.00\% & 43.67\%
      &                            & 61.00\% &                            \\
      \Xhline{1.2pt}

      \hline

      \multirow{15}{*}{\makecell[l]{\textbf{MOTIF} \\ \textbf{only stage3}}}
      & \multirow{3}{*}{1}  & Panda     & \cellcolor[HTML]{FEFCD6}80.00\% & \cellcolor[HTML]{FEFCD6}8.00\% & 96.00\% & 16.00\% & 88.00\% & 52.00\% & 56.67\%
      & \multirow{3}{*}{51.56\%} & 44.00\% & \multirow{3}{*}{30.67\%} \\
      &                         & xArm6     & 72.00\% & 92.00\% & \cellcolor[HTML]{FEFCD6}34.00\% & \cellcolor[HTML]{FEFCD6}6.00\% & 98.00\% & 54.00\% & 59.33\%
      &                            & 20.00\% &                            \\
      &                         & WidowX AI & 74.00\% & 50.00\% & 46.00\% &  6.00\% & \cellcolor[HTML]{FEFCD6}54.00\% & \cellcolor[HTML]{FEFCD6}2.00\% & 38.67\%
      &                            & 28.00\% &                            \\
      \cline{2-13}

      & \multirow{3}{*}{3}  & Panda     & \cellcolor[HTML]{FEFCD6}68.00\% & \cellcolor[HTML]{FEFCD6}28.00\% & 74.00\% & 34.00\% & 76.00\% & 48.00\% & 54.67\%
      & \multirow{3}{*}{57.44\%} & 48.00\% & \multirow{3}{*}{43.67\%} \\
      &                         & xArm6     & 66.00\% & 98.00\% & \cellcolor[HTML]{FEFCD6}76.00\% & \cellcolor[HTML]{FEFCD6}14.00\% & 96.00\% & 54.00\% & 67.33\%
      &                            & 45.00\% &                            \\
      &                         & WidowX AI & 68.00\% & 60.00\% & 56.00\% & 42.00\% & \cellcolor[HTML]{FEFCD6}76.00\% & \cellcolor[HTML]{FEFCD6}0.00\% & 50.33\%
      &                            & 38.00\% &                            \\
      \cline{2-13}

      & \multirow{3}{*}{5}  & Panda     & \cellcolor[HTML]{FEFCD6}84.00\% & \cellcolor[HTML]{FEFCD6}28.00\% & 94.00\% & 10.00\% & 88.00\% & 46.00\% & 58.33\%
      & \multirow{3}{*}{54.33\%} & 56.00\% & \multirow{3}{*}{47.33\%} \\
      &                         & xArm6     & 74.00\% & 94.00\% & \cellcolor[HTML]{FEFCD6}56.00\% & \cellcolor[HTML]{FEFCD6}26.00\% & 90.00\% & 68.00\% & 68.00\%
      &                            & 41.00\% &                            \\
      &                         & WidowX AI & 52.00\% & 44.00\% & 30.00\% &  4.00\% & \cellcolor[HTML]{FEFCD6}86.00\% & \cellcolor[HTML]{FEFCD6}4.00\% & 36.67\%
      &                            & 45.00\% &                            \\
      \cline{2-13}

      & \multirow{3}{*}{10} & Panda     & \cellcolor[HTML]{FEFCD6}92.00\% & \cellcolor[HTML]{FEFCD6}40.00\% & 72.00\% & 18.00\% & 86.00\% & 54.00\% & 60.33\%
      & \multirow{3}{*}{61.11\%} & 66.00\% & \multirow{3}{*}{58.00\%} \\
      &                         & xArm6     & 78.00\% & 92.00\% & \cellcolor[HTML]{FEFCD6}70.00\% & \cellcolor[HTML]{FEFCD6}50.00\% & 98.00\% & 62.00\% & 75.00\%
      &                            & 60.00\% &                            \\
      &                         & WidowX AI & 68.00\% & 34.00\% & 60.00\% & 30.00\% & \cellcolor[HTML]{FEFCD6}94.00\% & \cellcolor[HTML]{FEFCD6}2.00\% & 48.00\%
      &                            & 48.00\% &                            \\
      \cline{2-13}

      & \multirow{3}{*}{50} & Panda     & \cellcolor[HTML]{FEFCD6}98.00\% & \cellcolor[HTML]{FEFCD6}72.00\% & 94.00\% & 48.00\% & 96.00\% & 56.00\% & 77.33\%
      & \multirow{3}{*}{66.67\%} & 85.00\% & \multirow{3}{*}{71.67\%} \\
      &                         & xArm6     & 76.00\% & 94.00\% & \cellcolor[HTML]{FEFCD6}96.00\% & \cellcolor[HTML]{FEFCD6}42.00\% & 94.00\% & 72.00\% & 79.00\%
      &                            & 69.00\% &                            \\
      &                         & WidowX AI & 72.00\% &  2.00\% & 60.00\% &  6.00\% & \cellcolor[HTML]{FEFCD6}100.00\% & \cellcolor[HTML]{FEFCD6}22.00\% & 43.67\%
      &                            & 61.00\% &                            \\
      \Xhline{1.2pt}
    \end{tabular}%
  }
  \vskip -0.1in
\end{center}

\section{Qualitative Visualization}
\label{sec:app_visualization}

We provide some simulation and real-world execution examples; please see \cref{fig:simulation_total_combined} and \cref{fig:real_world_total} for details.

\begin{figure*}[htbp]
    \centering
    \begin{subfigure}[b]{1.0\textwidth}
        \centering
        \includegraphics[width=\textwidth]{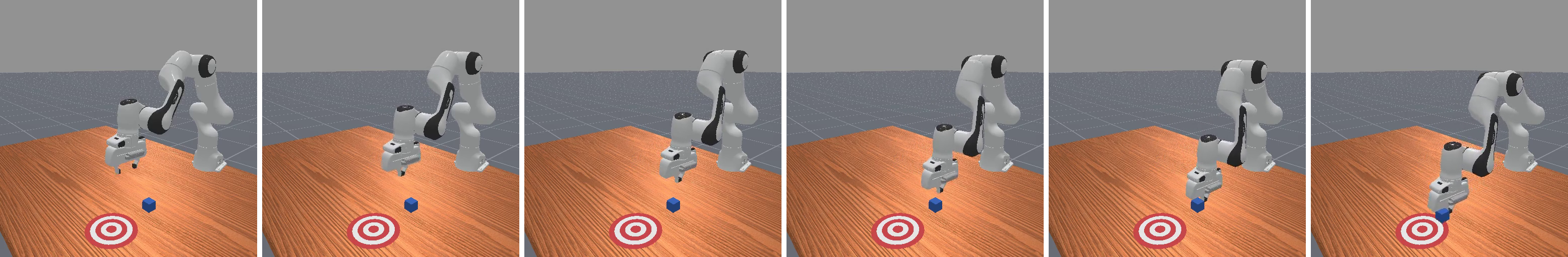}
        \caption{\textbf{Panda}: \textit{PushCube} (Push the cube to the target position.)}
        \label{fig:simulation_strip1}
    \end{subfigure}
    \vspace{2pt}

    \begin{subfigure}[b]{1.0\textwidth}
        \centering
        \includegraphics[width=\textwidth]{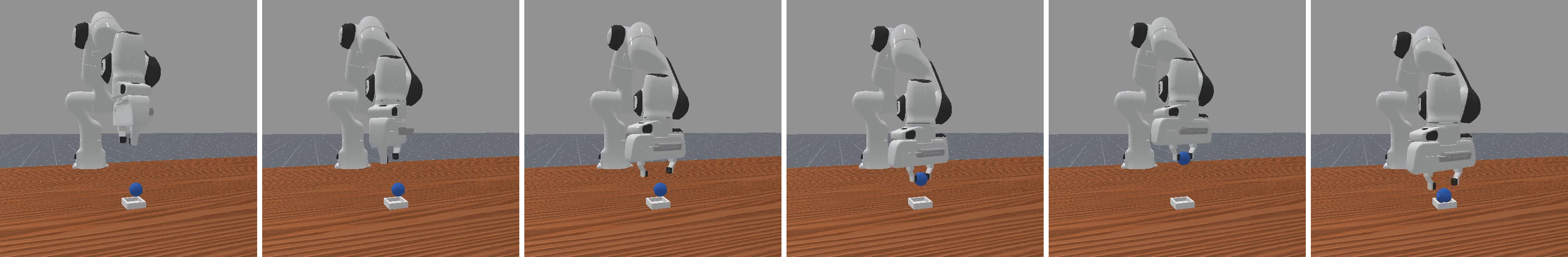}
        \caption{\textbf{Panda}: \textit{PlaceSphere} (Pick up the ball and place it in the target position.)}
        \label{fig:simulation_strip2}
    \end{subfigure}
    \caption{\textbf{Qualitative Simulation Results (Part I).} Visualization of the \textbf{Panda} arm executing the few-shot target tasks \textit{PushCube} and \textit{PlaceSphere}.} 
\end{figure*}

\begin{figure*}[htbp]
    \centering
    \ContinuedFloat 
    \begin{subfigure}[b]{1.0\textwidth}
        \centering
        \includegraphics[width=\textwidth]{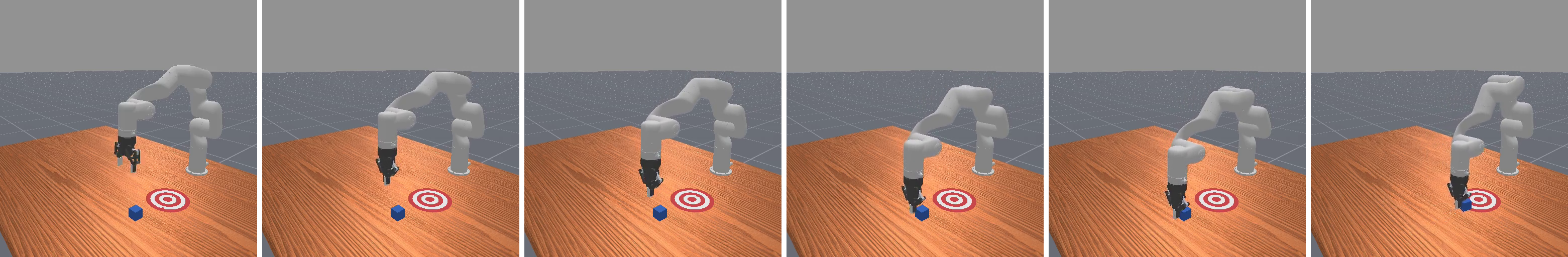}
        \caption{\textbf{xArm6}: \textit{PullCube} (Pull the cube to the target position.)}
        \label{fig:simulation_strip3}
    \end{subfigure}
    \vspace{2pt}

    \begin{subfigure}[b]{1.0\textwidth}
        \centering
        \includegraphics[width=\textwidth]{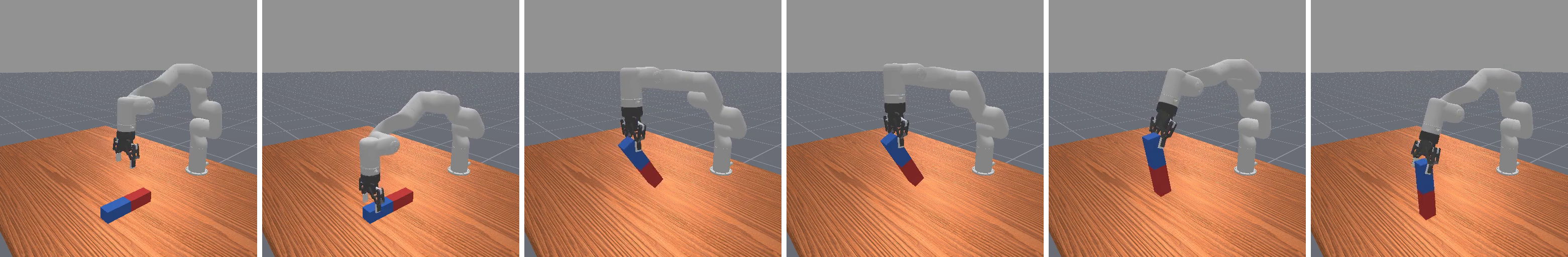}
        \caption{\textbf{xArm6}: \textit{LiftPegUpright} (Pick up the peg and place it upright.)}
        \label{fig:simulation_strip4}
    \end{subfigure}
    \caption{\textbf{Qualitative Simulation Results (Part II).} Visualization of the \textbf{xArm6} arm executing the few-shot target tasks \textit{PullCube} and \textit{LiftPegUpright}.} 
\end{figure*}

\begin{figure*}[htbp]
    \centering
    \ContinuedFloat 
    \begin{subfigure}[b]{1.0\textwidth}
        \centering
        \includegraphics[width=\textwidth]{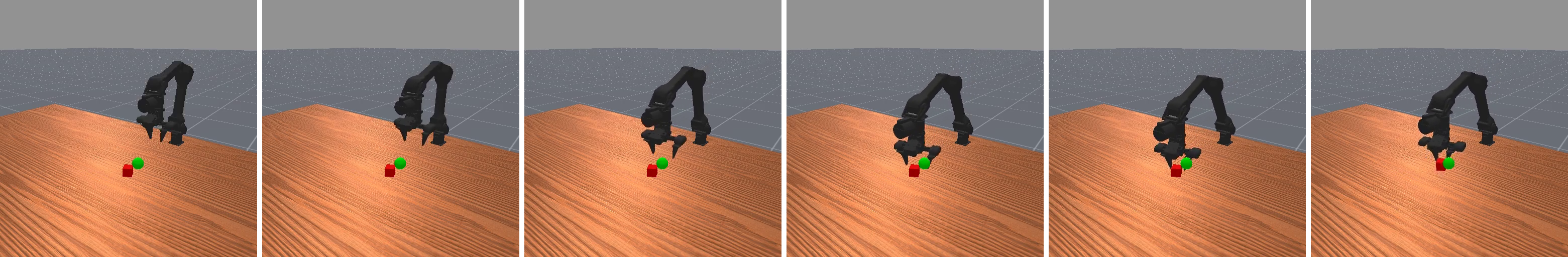}
        \caption{\textbf{WidowX AI}: \textit{PickCube} (Pick up the cube.)}
        \label{fig:simulation_strip5}
    \end{subfigure}
    \vspace{2pt}

    \begin{subfigure}[b]{1.0\textwidth}
        \centering
        \includegraphics[width=\textwidth]{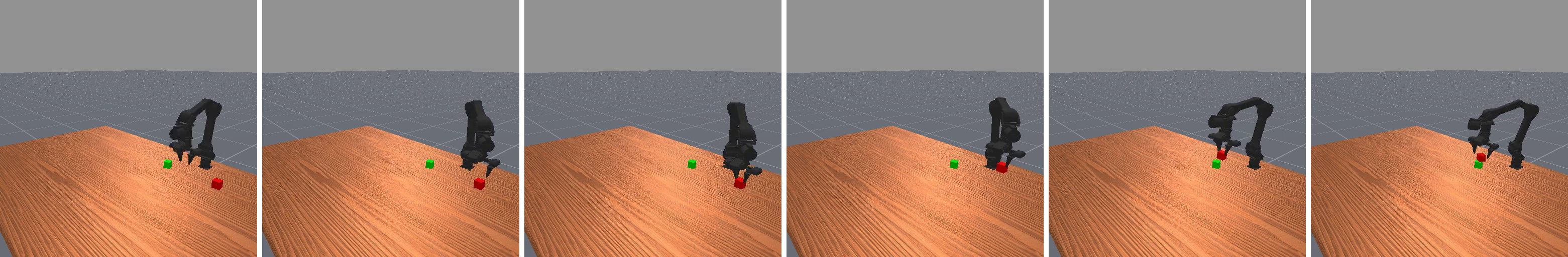}
        \caption{\textbf{WidowX AI}: \textit{StackCube} (Stack the cube on top of the other cube.)}
        \label{fig:simulation_strip6}
    \end{subfigure}
    \caption{\textbf{Qualitative Simulation Results (Part III).} Visualization of the \textbf{WidowX AI} arm executing the few-shot target tasks \textit{PickCube} and \textit{StackCube}.} 
    \label{fig:simulation_total_combined} 
\end{figure*}

\begin{figure*}[t]
    \centering
    \begin{subfigure}[b]{1.0\textwidth}
        \centering
        \includegraphics[width=\textwidth]{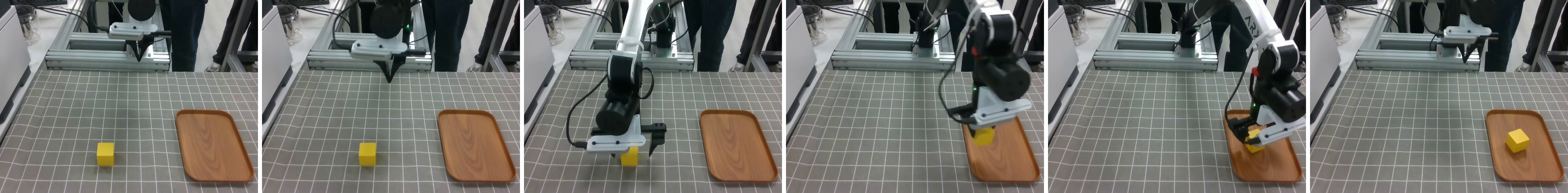}
        \caption{\textbf{ARX5}: \textit{PickPlace} (Pick up the cube and place on the plate.)}
        \label{fig:real_arx5_1}
    \end{subfigure}  
    \vspace{2pt} 

    \begin{subfigure}[b]{1.0\textwidth}
        \centering
        \includegraphics[width=\textwidth]{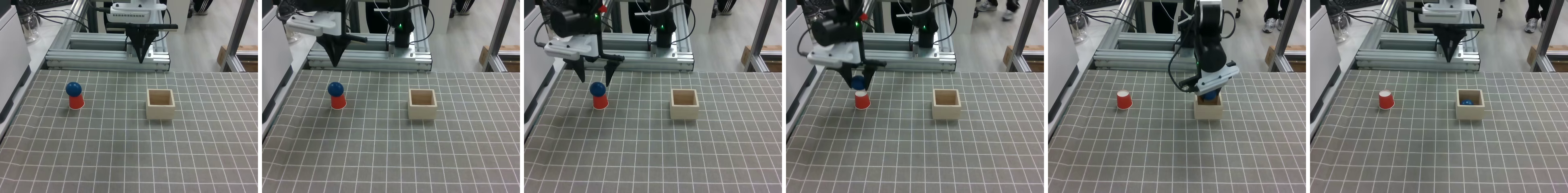}
        \caption{\textbf{ARX5}: \textit{PlaceSphere} (Pick up the ball and place into the box.)}
        \label{fig:real_arx5_2}
    \end{subfigure}
    \vspace{2pt}

    \begin{subfigure}[b]{1.0\textwidth}
        \centering
        \includegraphics[width=\textwidth]{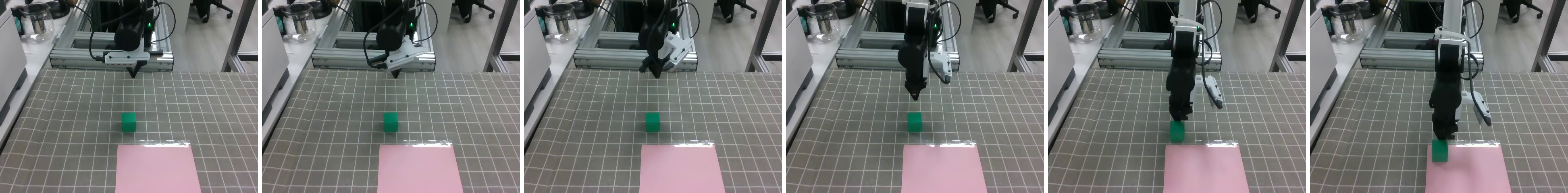}
        \caption{\textbf{ARX5}: \textit{PushCube} (Push the yellow cube into the pink area.)}
        \label{fig:real_arx5_3}
    \end{subfigure}
    \vspace{2pt}

    \begin{subfigure}[b]{1.0\textwidth}
        \centering
        \includegraphics[width=\textwidth]{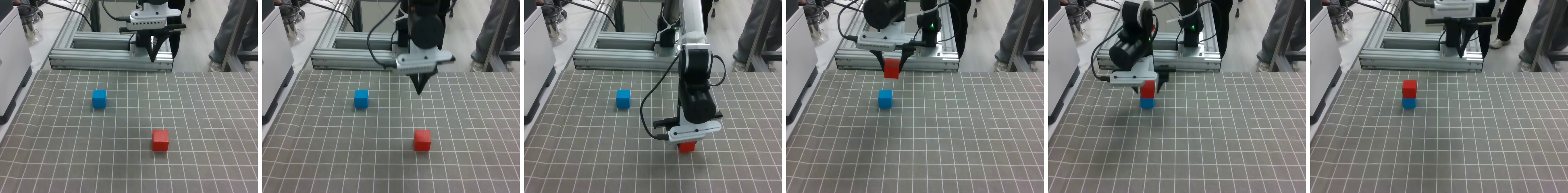}
        \caption{\textbf{ARX5}: \textit{StackCube} (Stack the red cube on the blue cube.)}
        \label{fig:real_arx5_4}
    \end{subfigure}

    \caption{\textbf{Qualitative Real-world Results (Part I).} Visualization of the \textbf{ARX5} embodiment executing four target tasks: \textit{PickPlace}, \textit{PlaceSphere}, \textit{PushCube}, and \textit{StackCube}.} 
\end{figure*}

\begin{figure*}[t]
    \centering
    \ContinuedFloat 
    \begin{subfigure}[b]{1.0\textwidth}
        \centering
        \includegraphics[width=\textwidth]{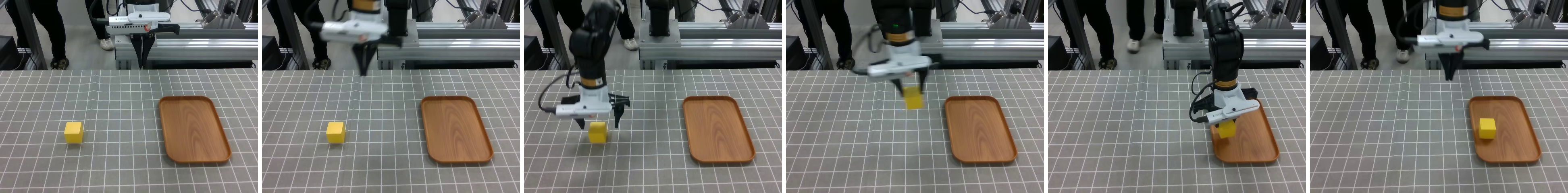}
        \caption{\textbf{Piper}: \textit{PickPlace} (Pick up the cube and place on the plate.)}
        \label{fig:real_piper_1}
    \end{subfigure}
    \vspace{2pt}

    \begin{subfigure}[b]{1.0\textwidth}
        \centering
        \includegraphics[width=\textwidth]{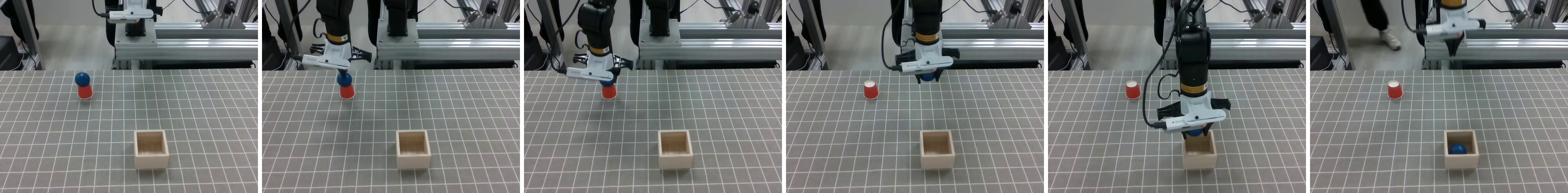}
        \caption{\textbf{Piper}: \textit{PlaceSphere} (Pick up the ball and place into the box.)}
        \label{fig:real_piper_2}
    \end{subfigure}
    \vspace{2pt}

    \begin{subfigure}[b]{1.0\textwidth}
        \centering
        \includegraphics[width=\textwidth]{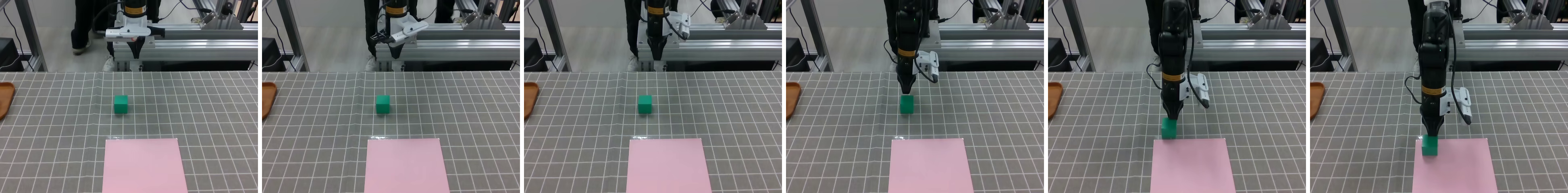}
        \caption{\textbf{Piper}: \textit{PushCube} (Push the yellow cube into the pink area.)}
        \label{fig:real_piper_3}
    \end{subfigure}
    \vspace{2pt}

    \begin{subfigure}[b]{1.0\textwidth}
        \centering
        \includegraphics[width=\textwidth]{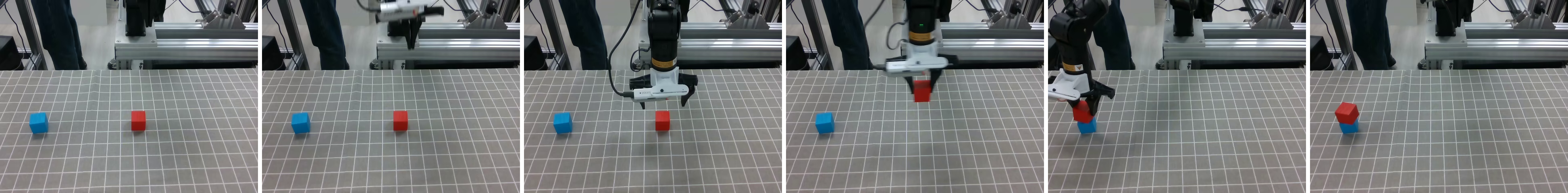}
        \caption{\textbf{Piper}: \textit{StackCube} (Stack the red cube on the blue cube.)}
        \label{fig:real_piper_4}
    \end{subfigure}

    \caption{\textbf{Qualitative Real-world Results (Part II).} Visualization of the \textbf{Piper} embodiment executing four target tasks: \textit{PickPlace}, \textit{PlaceSphere}, \textit{PushCube}, and \textit{StackCube}.} 
    \label{fig:real_world_total} 
\end{figure*}


\end{document}